%% file: 0main.tex
\documentclass{article}

\usepackage{microtype}
\usepackage{graphicx}
\usepackage{subfigure}
\usepackage{booktabs} 
\usepackage{kotex}
\usepackage{duckuments}

\usepackage{hyperref}

\usepackage[accepted]{icml2023}

\usepackage{amsmath}
\usepackage{amssymb}
\usepackage{mathtools}
\usepackage{amsthm}

\usepackage[capitalize,noabbrev]{cleveref}

\theoremstyle{plain}

\theoremstyle{definition}

\theoremstyle{remark}

\usepackage[textsize=tiny]{todonotes}

\input{macros}

\icmltitlerunning{Unsupervised Discovery of Semantic Latent Directions in Diffusion Models}

\begin{document}

\twocolumn[

\icmltitle{
Unsupervised Discovery of Semantic Latent Directions in Diffusion Models}




\icmlsetsymbol{equal}{*}

\begin{icmlauthorlist}
\icmlauthor{Yong-Hyun Park}{equal,snu}
\icmlauthor{Mingi Kwon}{equal,yonsei}
\icmlauthor{Junghyo Jo}{snu}
\icmlauthor{Youngjung Uh}{yonsei}

\end{icmlauthorlist}

\icmlaffiliation{snu}{Department of Physics Education, Seoul National University, Seoul, Korea}
\icmlaffiliation{yonsei}{Department of Artificial Intelligence, Yonsei University, Seoul, Korea}

\icmlauthorforemail{Yong-Hyun Park}{enkeejunior1@snu.ac.kr}
\icmlauthorforemail{Mingi Kwon}{kwonmingi@yonsei.ac.kr}

\icmlcorrespondingauthor{Youngjung Uh}{yj.uh@yonsei.ac.kr}
\icmlcorrespondingauthor{Junghyo Jo}{jojunghyo@snu.ac.kr}

\icmlkeywords{Machine Learning, Diffusion Model, Latent Space}

\vskip 0.3in
]



\OURSprintAffiliationsAndNotice{\icmlEqualContribution} 


\begin{abstract}

Despite the success of diffusion models (DMs), we still lack a thorough understanding of their latent space. While image editing with GANs builds upon latent space, DMs rely on editing the conditions such as text prompts.
We present an unsupervised method to discover interpretable editing directions for the latent variables $\vx_t\in\mathcal{X}$ of DMs. Our method adopts Riemannian geometry between \exspace{} and the intermediate feature maps \ehspace{} of the U-Nets to provide a deep understanding over the geometrical structure of \exspace{}. The discovered semantic latent directions mostly yield disentangled attribute changes, and they are globally consistent across different samples. Furthermore, editing in earlier timesteps edits coarse attributes, while ones in later timesteps focus on high-frequency details. We define the curvedness of a line segment between samples to show that \exspace{} is a curved manifold. Experiments on different baselines and datasets demonstrate the effectiveness of our method even on Stable Diffusion. Our source code will be publicly available for the future researchers.

\end{abstract}

\input{1intro.tex}

\input{2relatedwork.tex}

\input{3method.tex}

\input{4experiment.tex}
\input{5conclusion.tex}
\nocite{langley00}

\bibliography{example_paper}
\bibliographystyle{icml2023}

\newpage
\input{6appendix.tex}



\end{document}

%% file: macros.tex
\usepackage{amsmath,amsfonts,bm}

\newcommand{\fref}[1]{Figure~\ref{#1}}
\newcommand{\eref}[1]{Eq.~(\ref{#1})}
\newcommand{\tref}[1]{Table~\ref{#1}}
\newcommand{\sref}[1]{$\S$~\ref{#1}}
\newcommand{\aref}[1]{Appendix \ref{#1}}

\newcommand{\ehspace}{$\mathcal{H}$}
\newcommand{\exspace}{$\mathcal{X}$}

\newcommand\mingi[1]{\textcolor{orange}{#1}}

\def\vepsilon{\bm{\epsilon}}
\def\tepsilont{\vepsilon^{\theta}_{t}}

\def\jacx{{J}_{\mathbf{x}}}
\def\tanxspace{\mathcal{T}_\mathbf{x}}
\def\tanhspace{\mathcal{T}_\mathbf{h}}

\def\vx{\mathbf{x}}
\def\vh{\mathbf{h}}
\def\vz{\mathbf{z}}
\def\vv{\mathbf{v}}
\def\dx{\mathbf{v}}
\def\dh{\mathbf{u}}
\def\predictedx0{x_0}
\newcommand*{\tran}{^{\mkern-1.5mu\mathsf{T}}}



%% file: 1intro.tex
\section{Introduction}
Diffusion models (DMs) are highly powerful generative models that have shown great performance \cite{ho2020denoising,song2020denoising,song2020score,dhariwal2021diffusion,nichol2021improved}.
To control the generative process, existing methods have introduced conditional DMs, especially for text-to-image synthesis \cite{ramesh2022hierarchical,rombach2022high,balaji2022ediffi,nichol2021glide}, or mixing the latent variables $\vx_t$ of different sampling processes \cite{choi2021ilvr,meng2021sdedit,avrahami2022blended,liew2022magicmix,kawar2022imagic,avrahami2022spatext}.

Despite their success, the research community still lacks a clear understanding of what the latent variables or intermediate features of the models are embedded or how they are reflected in the resulting images. We attribute it to the characteristic iterative process of the DMs which involves a sequence of noisy images and subtle noises, i.e., the embeddings are not directly connected to the final images.
In contrast, arithmetic operations in the latent space of generative adversarial networks (GANs) lead to semantic changes in the resulting images \cite{goodfellow2020generative}. This property has been one of the key factors in developing GANs for real-world applications. We suppose that a better understanding of the latent space of DMs will boost similar development.

\begin{figure}[!t]
    \centering
    \includegraphics[width=0.9\linewidth]{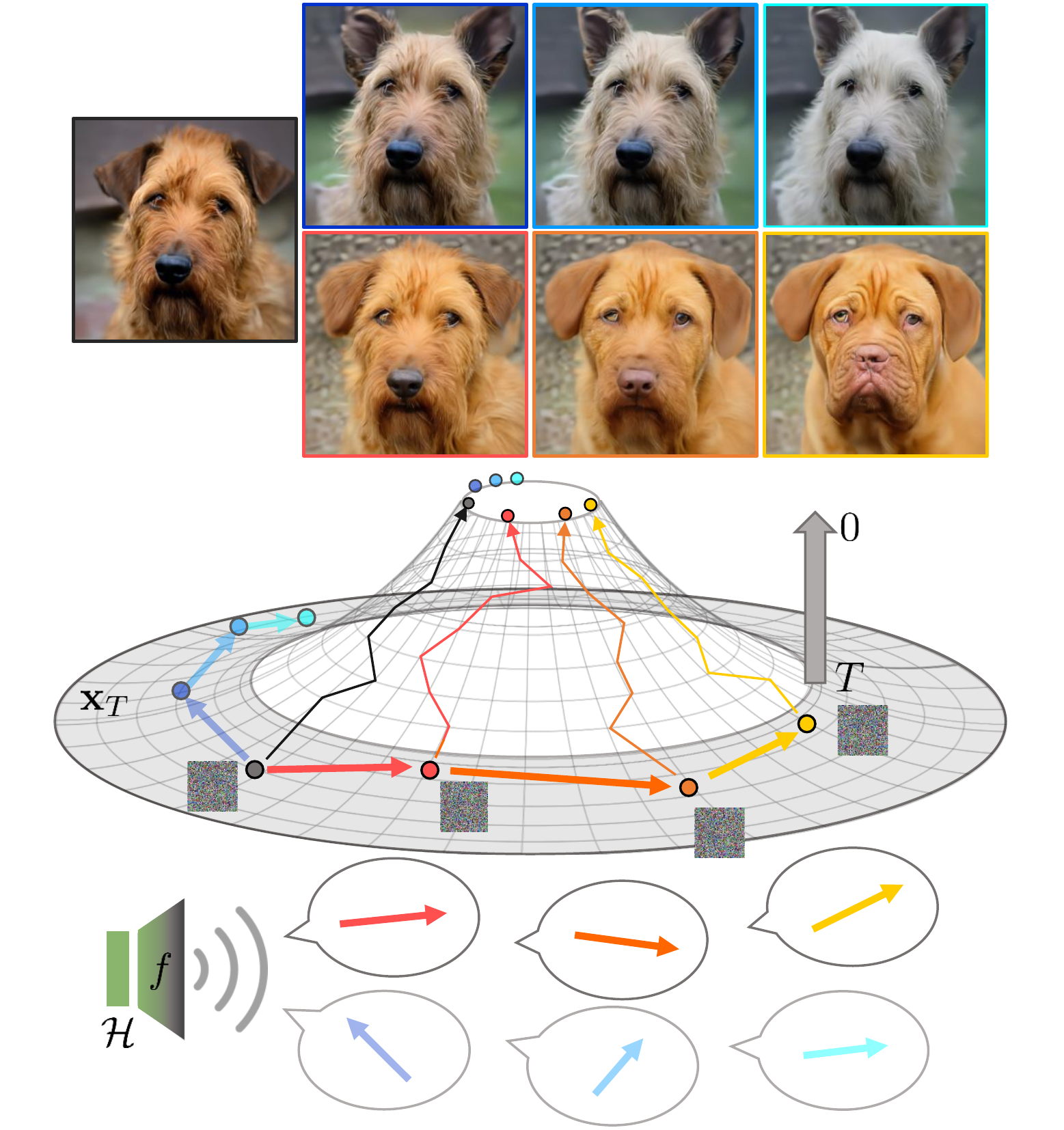}
    \caption{\textbf{Conceptual illustration of our method.} 
    We find semantic directions in the latent space $\vx_t$ in an unsupervised fashion relying on Riemannian geometry between $\vx_t$ and $\vh_t$. $\mathcal{H}$ denotes the bottleneck layer and $f$ indicates the frozen encoder of a U-Net. 
    The found directions manipulate the semantics of the resulting images.
    }
    \vspace{-1em}
    \label{fig:main_concept}
\end{figure}

\citet{kwon2022diffusion} adopt the intermediate feature space of the diffusion kernel as a semantic latent space, namely \ehspace{}, paired with a designated asymmetric sampling process. They revealed the local linearity of \ehspace{}, adding to our understanding of the latent space of DMs.
However, they do not directly deal with the latent variables $\vx_t$ but rely only on a proxy, $\mathbf{h}$. Furthermore, they require external supervision such as Contrastive Language-Image Pretraining (CLIP) to find editable directions. \cite{radford2021learning}

In this paper, we introduce useful intuitions about latent space \exspace{} to deepen our understanding of how we can control \emph{pretrained and frozen} diffusion models. First, we identify semantic latent directions in \exspace{} which manipulate the resulting images using Riemannian geometry in an unsupervised manner. The directions come from the singular value decomposition of the Jacobian of the mapping from \exspace{} to \ehspace{}, the intermediate feature space of the model. \fref{fig:main_concept} illustrates the main concept of our method.

Second, we find global semantic directions by exploiting the homogeneity of \ehspace{}. It removes cumbersome per-sample Jacobian computation and allows general controllability. It follows the course of generative adversarial networks: extending per-sample editing directions \cite{ramesh2018spectral,patashnik2021styleclip,abdal2021styleflow,shen2021closed} to global editing directions \cite{harkonen2020ganspace,shen2021closed,yuksel2021latentclr}.


Last but not least, we show interesting properties of the diffusion models. Spherical linear interpolation in \exspace{} leads to smooth interpolation between samples because it is approximately geodesic in \ehspace{}. That is, \exspace{} is a warped space. The early timesteps generate low frequency components and the later timesteps generate high frequency components. Although it is indirectly shown in existing works \citet{choi2022perception}, we explicitly reveal it via power spectral density.

In the experiments, we demonstrate that the directions found in an unsupervised manner indeed lead to semantic changes in the images. We note that discovering the editing directions in the latent variables of diffusion models has not been tackled. Furthermore, we provide thorough quantitative and qualitative analyses on the aforementioned properties. Our method even works on stable diffusion \cite{rombach2022high}.




%% file: 2relatedwork.tex
\section{Related Works}
Recent advances in DMs have resulted in the development of a universal approach known as DDPMs \cite{ho2020denoising}. \citet{song2020score} have facilitated the unification of DMs with score-based models using SDEs. However, further studies still remain to fully understand and utilize the capabilities of DMs.

An important subject is the introduction of gradient guidance, including classifier-free guidance, to control the generative process \cite{dhariwal2021diffusion,sehwag2022generating,avrahami2022blended,liu2021more,nichol2021glide,rombach2022high}. \citet{choi2021ilvr} and \citet{meng2021sdedit} have attempted to manipulate the resulting images of DMs by replacing latent variables, allowing the generation of desired random images. However, due to the lack of semantics in the latent variables of DMs, current approaches have critical problems with semantic image editing.

Alternative approaches have explored the potential of using the feature space within the U-Net for semantic image manipulation. For example, \citet{baranchuk2021label} and \citet{tumanyan2022plug} use the feature map of the U-Net for semantic segmentation and maintaining the structure of generated images. \citet{kwon2022diffusion} have shown that the bottleneck of the U-Net can be used as a semantic latent space.
The experimental observation lacks a theoretical understanding of the feature map of DMs.

The study of latent spaces has gained significant attention in recent years. In the field of Generative Adversarial Networks (GANs), researchers have proposed various methods to manipulate the latent space to achieve the desired effect in the generated images. For example, local latent space manipulation techniques such as \cite{ramesh2018spectral,patashnik2021styleclip,abdal2021styleflow} have been developed, as well as global manipulation techniques such as \cite{harkonen2020ganspace,shen2021closed,yuksel2021latentclr}. More recently, several studies \cite{zhu2021low, choi2021not} have examined the geometrical properties of latent space in GANs and utilized these findings for image manipulations. These studies bring the advantage of better understanding the characteristics of the latent space and facilitating the analysis and utilization of GANs. In contrast, the latent space of DMs remains poorly understood, making it difficult to fully utilize their capabilities.


Some studies have applied Riemannian geometry to analyze the latent spaces of deep generative models, such as Variational Autoencoders (VAEs) and GANs. \cite{arvanitidis2017latent, shao2018riemannian, chen2018metrics, arvanitidis2020geometrically} \citet{shao2018riemannian} proposed a pullback metric on the latent space from image space Euclidean metric to analyze the latent space's geometry. This method has been widely used in VAEs and GANs because it only requires a differentiable map from latent space to image space.
However, it has limitations such as a lack of evidence for applying the Euclidean metric in image space and the absence of a global semantic direction for manipulating arbitrary samples.
Moreover, no studies have investigated the geometry of latent space of DMs utilizing the pullback metric.

%% file: 3method.tex

\section{Editing with semantic latent directions}

This section explains how we extract the interpretable directions in the latent space of DMs using differential geometry. 
First, we adopt the local Euclidean metric of \ehspace{} to identify semantic directions for individual samples in \exspace{}.
Second, we find global semantic directions by averaging the local semantic directions of individual samples. Then, we use the global directions to manipulate any sample to have the same interpretable features.
Finally, we introduce a normalization technique to prevent distortion.




\begin{figure}[!t]
    \centering\includegraphics[width=0.8\linewidth]{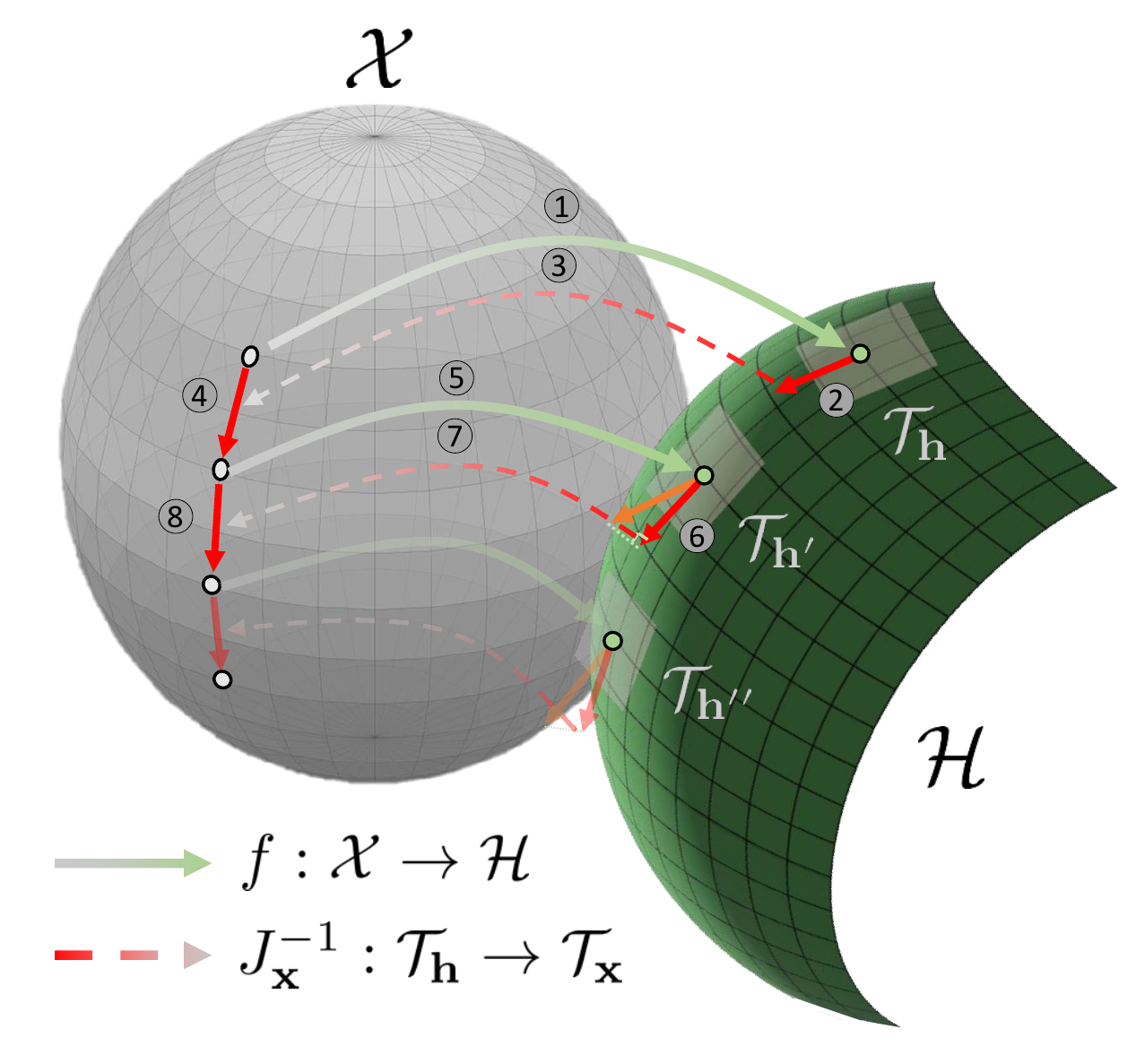}
    \caption{
    \textbf{Conceptual illustration of our editing procedure.} 
    It consists of two parts: (\textcircled{1} $\sim$ \textcircled{4}) discovering semantic latent directions using pullback metric and (\textcircled{5} $\sim$ \textcircled{8}) editing samples multiple times through geodesic shooting.
    \textcircled{1} Map a sample in $\mathcal{X}$ into a tangent space $\mathcal{T}_\mathbf{h}$ in $\mathcal{H}$. 
    \textcircled{2} Choose a direction in $\mathcal{T}_\mathbf{h}$.
    \textcircled{3} Find its corresponding direction in $\mathcal{X}$ using $\jacx{}^{-1}$.
    \textcircled{4} Edit the sample by adding the discovered direction after normalizing to a predefined length.
    \textcircled{5} Map the edited sample to a new tangent space $\mathcal{T}_{\mathbf{h}'}$ in $\mathcal{H}$ for multiple editing.
    \textcircled{6} Using parallel transport, move the direction chosen in \textcircled{2} to the new tangent space $\mathcal{T}_{\mathbf{h}'}$.
    \textcircled{7}-\textcircled{8} Repeat \textcircled{3}-\textcircled{4}. And then repeat \textcircled{5}-\textcircled{8}.
}
    \vspace{-1em}
    \label{fig:method_figure}
\end{figure}

\subsection{Pullback metric}
We consider a curved manifold, $\mathcal{X}$, where our latent variables $\mathbf{x}_t$ exist. 
The differential geometry represents $\mathcal{X}$ through patches of tangent spaces, $\tanxspace{}$, which are vector spaces defined at each point $\mathbf{x}$. Then, all the geometrical properties of $\mathcal{X}$ can be obtained from the metric of $||d\mathbf{x}||^2 = \langle d{\mathbf{x}},d{\mathbf{x}} \rangle_\mathbf{x}$ in $\tanxspace{}$.
However, we do not have any knowledge of $\langle d{\mathbf{x}},d{\mathbf{x}} \rangle_\mathbf{x}$.
It is definitely not a Euclidean metric. Furthermore, samples of $\mathbf{x}_t$ at intermediate timesteps of DMs include inevitable noise, which prevents finding semantic directions in $\tanxspace{}$.

Fortunately, \citet{kwon2022diffusion} observed that \ehspace{}, defined by the bottleneck layer of the U-Net, exhibits local linearity.
This allows us to adopt the Euclidean metric on $\mathcal{H}$.
In differential geometry, when a metric is not available on a space, {\it pullback metric} is used.
If a smooth map exists between the original metric-unavailable space and a metric-available space, the pullback metric of the mapped space is used to measure the distances in the original space.   
Our idea is to use the pullback Euclidean metric on $\mathcal{H}$ to define the distances between the samples in $\mathcal{X}$.

DMs are trained to infer the noise $\mathbf{\epsilon}_t$ from a latent variable $\mathbf{x}_t$ at each diffusion timestep $t$. 
Each $\mathbf{x}_t$ has a different internal representation $\mathbf{h}_t$, the bottleneck representation of the U-Net, at different $t$'s.
The differentiable map between $\mathcal{X}$ and $\mathcal{H}$ is denoted as $f : \mathcal{X} \rightarrow \mathcal{H}$.
Hereafter, we refer to $\mathbf{x}_t$ as $\mathbf{x}$ for brevity unless it causes confusion. It is important to note that our method can be applied at any timestep in the denoising process.
The differential geometry then defines a linear map between the tangent space $\tanxspace{}$ at $\mathbf{x}$ and corresponding tangent space $\tanhspace{}$ at $\mathbf{h}$. The linear map can be described by the {\it Jacobian} $J_{\mathbf{x}} = \nabla_{\mathbf{x}} \mathbf{h}$
which determines how a vector $\mathbf{v} \in \tanxspace{}$ is mapped into a vector $\mathbf{u} \in \tanhspace{}$ by
$\mathbf{u} = J_{\mathbf{x}} \mathbf{v}$. 
In practice, the Jacobian can be computed from automatic differentiation of the U-Net.
However, since the Jacobian of too many parameters is not tractable, we use a sum-pooled feature map of the bottleneck representation as our $\mathcal{H}$. 

Using the local linearity of $\mathcal{H}$, we assume the metric, $||d\mathbf{h}||^2 = \langle d\mathbf{h}, d\mathbf{h} \rangle_\mathbf{h} = d\mathbf{h}^{\tran} d\mathbf{h}$ as a usual dot product defined in the Euclidean space.
To assign a geometric structure to $\mathcal{X}$, we use the pullback metric of the corresponding $\mathcal{H}$.
The pullback norm of $\mathbf{v} \in \tanxspace{}$ is defined as follows:
\begin{equation}
\begin{aligned}
||\dx{}||^2_\text{pb} \triangleq  \langle \dh{}, \dh{} \rangle_{\mathbf{h}} = \dx{}^{\tran} \jacx{}^{\tran} \jacx{}\dx{}.
\end{aligned}
\end{equation}

\subsection{Extracting the semantic directions and editing}
\label{sec:method_local}
This subsection describes how we extract semantic latent directions using the pullback metric, and how we edit samples for multiple times given the meaningful directions by geodesic shooting. The overall process is illustrated in \fref{fig:method_figure}.

\paragraph{Semantic latent directions}
Using the pullback metric, we can extract semantic directions of $\mathbf{v} \in \tanxspace{}$ that show large variability of the corresponding $\mathbf{u} \in \tanhspace{}$. 
We find a unit vector $\mathbf{v}_1$ that maximizes $||\dx{}||^2_\text{pb}$.
In practice, $\mathbf{v}_1$ corresponds to the first right singular vector from the singular value decomposition of $\jacx{} = U \Lambda V^{\tran}$. It can be interpreted as the first eigenvector of $\jacx{}^{\tran} \jacx{} = V \Lambda^2 V^{\top}$.
By maximizing $||\dx{}||^2_\text{pb}$ while remaining orthogonal to $\mathbf{v}_1$, one can obtain the second unit vector $\mathbf{v}_2$. This process can be repeated to have $n$ semantic directions of $\{\mathbf{v}_1, \mathbf{v}_2, \cdots, \mathbf{v}_n \}$ in $\mathcal{T}_{\mathbf{x}}$.

Using the linear transformation between $\tanxspace{}$ and $\tanhspace{}$ via the Jacobian $\jacx{}$, one can also obtain semantic directions in $\mathcal{T}_{\mathbf{h}}$:
\begin{align}
    \dh{}_i = \frac{1}{\lambda_i} \jacx{} \dx{}_i.
\end{align}
Here, we normalize $\dh{}_i$ by dividing the $i$-th singular value $\lambda_i$ of $\Lambda$ to preserve the Euclidean norm $||\mathbf{u}_i||=1$. 
After selecting the top $n$ (e.g. $n = 50$) directions of large eigenvalues, we can approximate any vector in $\mathcal{T}_{\mathbf{h}}$ with finite basis, $\{\mathbf{u}_1, \mathbf{u}_2, \cdots, \mathbf{u}_n \}$.
When we refer to a tangent space henceforth, it means the $n$-dimensional low-rank approximation of the original tangent space.

\paragraph{Iterative editing with geodesic shooting} 
Now, we edit a sample with the $i$-th semantic direction through $\mathbf{x} \to \mathbf{x}' = \mathbf{x} + \gamma \mathbf{v}_i$, where $\gamma$ is a hyper-parameter 
that controls the size of the editing.
If we want to increase the editing strength, we need to repeat the same operation. However, this would not work because $\mathbf{v}_i$ may escape from the tangent space $\mathcal{T}_{\mathbf{x}'}$.
Thus, it is necessary to relocate the extracted direction to a new tangent space.
To achieve this, we use {\it parallel transport} that projects $\mathbf{v}_i$ onto the new tangent space $\mathcal{T}_{\mathbf{x}'}$. 
Parallel transport moves a vector without changing its direction as much as possible, while keeping the vector tangent on the manifold~\cite{shao2018riemannian}.
It is notable that the projection significantly modifies the original vector $\mathbf{v}_i$, because $\mathcal{X}$ is a curved manifold.
However, $\mathcal{H}$ is relatively flat. Therefore, it is beneficial to apply the parallel transport in $\mathcal{H}$.

To project $\mathbf{v}_i$ onto the new tangent space $\mathcal{T}_{\mathbf{x}'}$, we use parallel transport in $\mathcal{H}$. First, we convert the semantic direction $\mathbf{v}_i$ in $\mathcal{T}_{\mathbf{x}}$ to the corresponding direction of $\mathbf{u}_i$ in $\mathcal{T}_{\mathbf{h}}$. Second, we apply the parallel transport $\mathbf{u}_i \in \mathcal{T}_{\mathbf{h}}$ to ${\mathbf{u}'}_{i} \in \mathcal{T}_{\mathbf{h}'}$, where $\mathbf{h}' = f(\mathbf{x}')$. 
The parallel transport has two steps. The first step is to project $\mathbf{u}_i$ onto a new tangent space. This step keeps the vector tangent to the manifold. The second step is to normalize the length of the projected vector. This step preserves the size of the vector.
Third, we obtain $\mathbf{v}'_{i}$ by transforming $\mathbf{u}'_{i}$ into $\mathcal{X}$.
Using this parallel transport of $\mathbf{v}_i \to \mathbf{v}'_i$ via $\mathcal{H}$, we can realize the multiple feature editing of $\mathbf{x} \to \mathbf{x}' = \mathbf{x} + \gamma\mathbf{v}_i \to \mathbf{x}'' = \mathbf{x}' + \gamma \mathbf{v}'_i $.
Based on the definition of Jacobian, this editing process can be viewed as a movement in \ehspace{} with the corresponding direction, i.e., $\mathbf{h} \to \mathbf{h}' = \mathbf{h} + \delta \mathbf{u}_i \to \mathbf{h}'' = \mathbf{h}' + \delta \mathbf{u}'_i$. 
This iterative editing procedure is called {\it geodesic shooting}, since it naturally forms a geodesic ~\cite{shao2018riemannian}.
\fref{fig:method_figure} summarizes the above procedure. 
See \aref{appendixsec:algorithm} for details.

\begin{figure}[!t]
    \centering
    \includegraphics[width=1\linewidth]{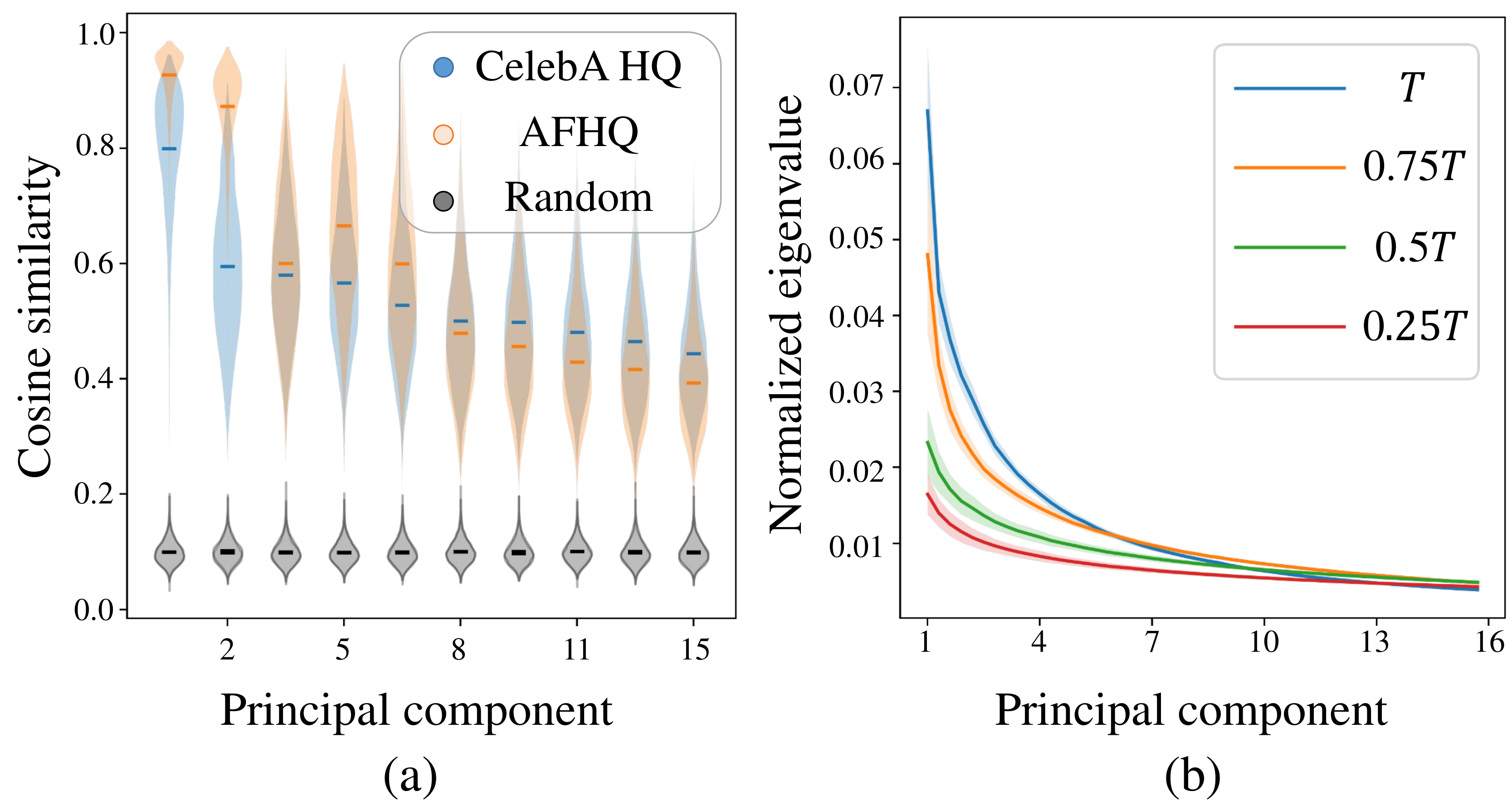}
    \caption{
    \textbf{(a) Homogeneity across local directions from different images.} A distribution represents the statistics of maximum cosine similarities between the principal directions of pairs of 100 samples in $\mathbf{x}_T$.
    The top principal directions better align than the rest.
    The comparison with random directions (black) confirms that the similarity does not arise by chance.
    \textbf{(b) Eigenvalue spectrum of $\jacx{}$ at different timesteps.}
    Early timesteps ($t\approx T$) have larger top eigenvalues implying fewer but more eminent directions than later timesteps. 
    }
    \vspace{-1em}
    \label{fig:homogenity}
\end{figure}

\subsection{Global semantic directions}
\label{sec:method_global}
We extracted meaningful directions for editing $\mathbf{x}_t$.
However, the semantic latent directions are {\it local}, and thus are applicable only to individual samples of $\mathbf{x}_t$.
Thus, we need to obtain {\it global} semantic directions that have the same semantic meaning for every sample.
In this study, we observed a large overlap between the latent directions of individual samples.
This observation motivates us to hypothesize that \ehspace{} has global semantic directions.
To verify this hypothesis, we investigate whether, for any $\dh_i^{(1)} \in \mathcal{T}_{\mathbf{h}^{(1)}}$, there exists $\dh_j^{(2)} \in \mathcal{T}_{\mathbf{h}^{(2)}}$ that has a large overlap with $\dh_i^{(1)}$.
Then, we compare latent directions of $\dh_i^{(1)}$ between many samples of $\mathbf{x}_t$.
For the dominant directions of $\mathbf{u}_i^{(1)}$ and $\mathbf{u}_j^{(2)}$ with large eigenvalues of $\lambda_i^{(1)}$ and $\lambda_j^{(2)}$, we always found a good pair of $(i, j)$ that showed a significant overlap between the two unit vectors when $t = T$ (\fref{fig:homogenity} (a)).
Thus, we define global semantic directions, $\bar{\mathbf{u}}_i$, by averaging the closest latent directions in $\mathcal{H}$ of individual samples of $\mathbf{x}_T$. 
The global direction can be used to edit any sample $\mathbf{x}$.
Note that $\bar{\mathbf{u}}_i$ can sometimes escape from the local tangent space of $\tanhspace{}$. 
To mitigate this escape, we project $\bar{\mathbf{u}}_i$ into $\tanhspace{}$. Since our method edits the sample in $\mathcal{X}$, we transform $\bar{\mathbf{u}}_i$ into the corresponding direction $\bar{\mathbf{v}}_i$ in $\mathcal{T}_{\mathbf{x}}$ via the Jacobian.

However, it is cautious to apply our hypothesis when we consider $\mathbf{x}_t$ for small $t$.
We compared eigenvalue spectra between different $t$, and observed that they become flatter as $t$ is closer to 0 (\fref{fig:homogenity} (b)).
This shows that a few dominant feature directions exist for $\mathbf{x}_T$, whereas diverse feature directions exist for $\mathbf{x}_t$ with small $t$. 
Then, it is difficult to define global directions based on the homogeneity of local feature directions.





\subsection{Normalizing distortion due to editing}

DMs generate images by iteratively denoising $\mathbf{x}_T \to \mathbf{x}_{T-1} \to \cdots \to \mathbf{x}_0$.
Suppose that we edit an image of $\mathbf{x}_t$ at a time step $t$ with $\mathbf{x}_t \to \mathbf{x}_t + \gamma \mathbf{v}_i$.
The editing signal of $\mathbf{v}_i$ is propagated and amplified throughout the denoising process. 
The amplification may lead to unexpected artifacts in generating $\mathbf{x}_0$.
To avoid this problem, some normalization of $\mathbf{x}_t$ is necessary after the editing.
However, it is difficult to normalize only the signal inside $\mathbf{x}_t$ that is mixed with white noise.
Here, we propose an improved editing method.

\begin{figure}[!t]
    \centering
    \includegraphics[width=0.9\linewidth]{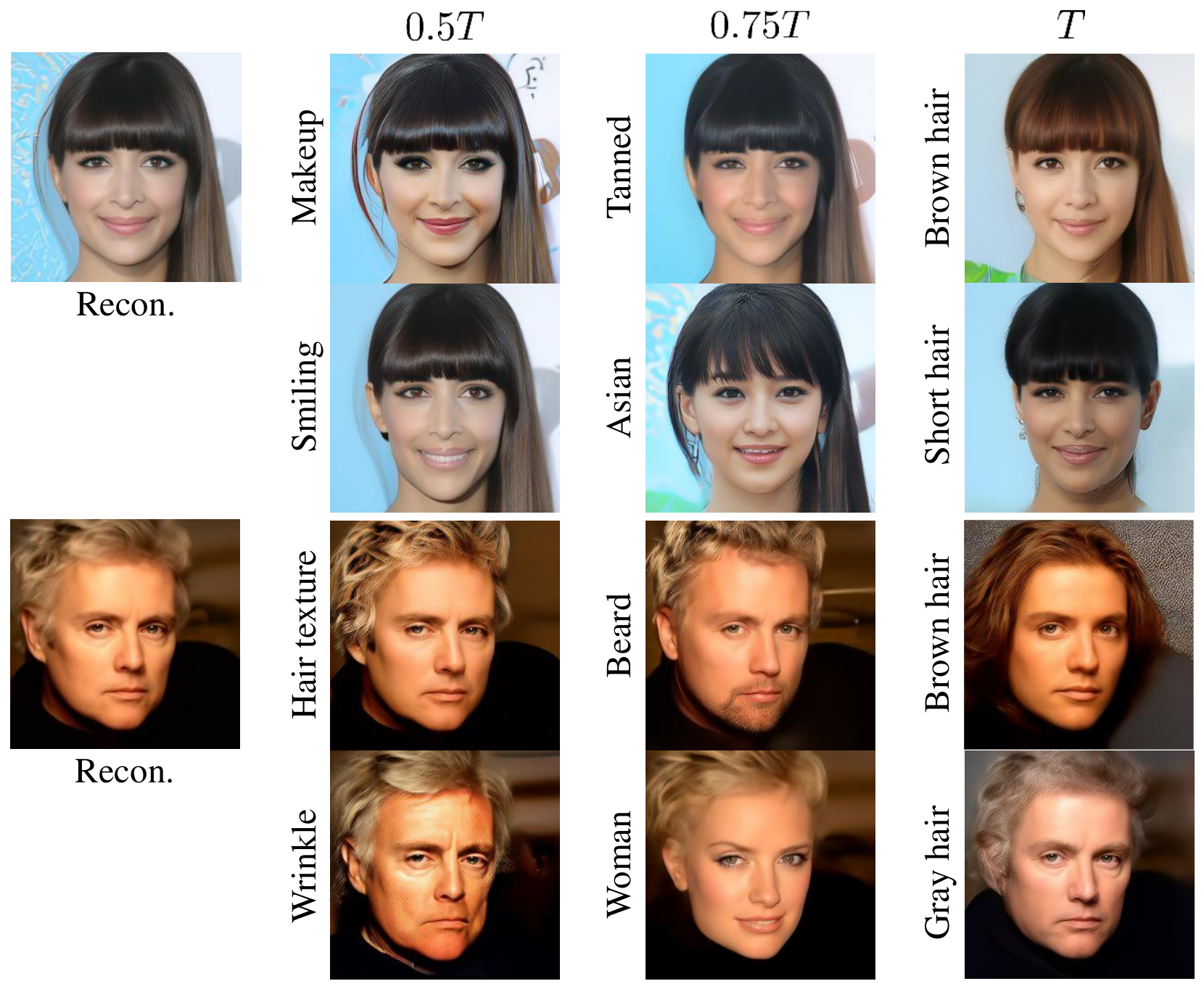}
    \centering
    \includegraphics[width=0.9\linewidth]{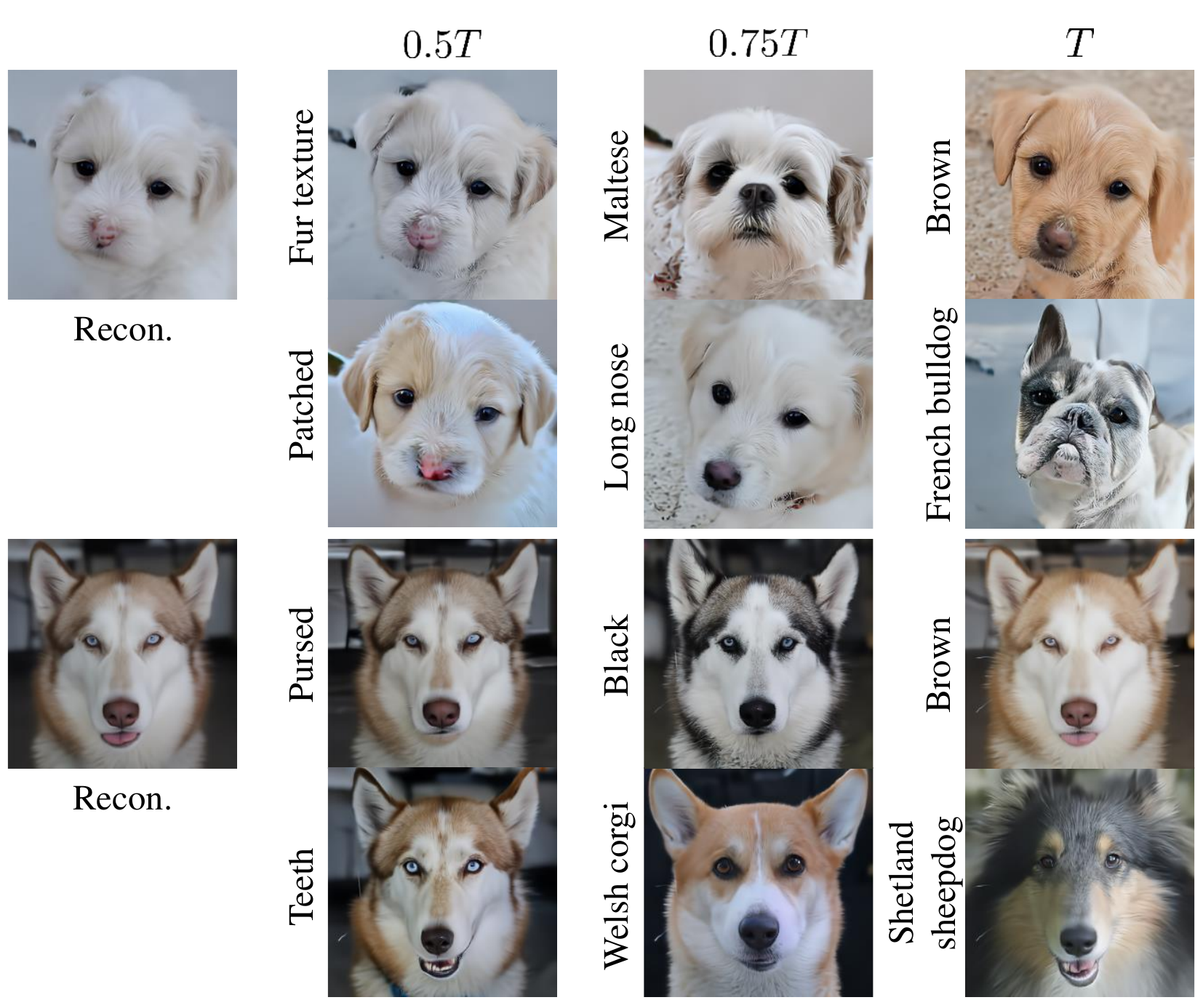}
    \caption{
    \textbf{Example edited results by the semantic latent directions.} ``Recon." denotes the reconstruction of real images through DDIM inversion. The attributes are manually interpreted because the directions are not supervised. Different columns are edited at different denoising timesteps ($0.5T$, $0.75T$, and $T$).
    }
    \label{fig:local_basis}
\end{figure}

Denoising diffusion implicit models (DDIM) computes $\mathbf{x}_0$ from $\mathbf{x}_t$ with predicted noise $\tepsilont (\mathbf{x}_t)$~\cite{song2020denoising}:
\begin{equation}
\sqrt{\alpha_t} \mathbf{x}_0 = \mathbf{x}_t - \sqrt{1-\alpha_t}\tepsilont(\mathbf{x}_t).
\end{equation}
With a little abuse of notation, let $\mathbf{x}_0(\mathbf{x}_t)$ be a function of $\mathbf{x}_t$. 
In an ideal scenario, $\mathbf{x}_0(\mathbf{x}_t)$ can be assumed to contain only the signal of $\mathbf{x}_t$, which simplifies the regularization process~\cite{zhang2022gddim}. 
Our improved editing method consists of three steps.
First, we edit the original image as $\mathbf{x}_t \to \mathbf{x}_t + \gamma \mathbf{v}_i$. Second, we regularize $\mathbf{x}_0(\mathbf{x}_t+\gamma 
 \dx{}_i)$ to preserve its signal after the edition. Regularization is implemented by normalizing the pixel-to-pixel standard deviation of $\mathbf{x}_0(\mathbf{x}_t+\gamma \dx{}_i)$, while keeping it's mean pixel values fixed.
We denote the normalized $\mathbf{x}_0(\mathbf{x}_t + \gamma \mathbf{v}_i)$ as $\mathbf{x}'_0$.
Third, we solve the DDIM equation for $\mathbf{x}'_t$, $\sqrt{\alpha_t} \mathbf{x}'_0 = \mathbf{x}'_t - \sqrt{1-\alpha_t}\tepsilont(\mathbf{x}'_t)$, to obtain a corresponding edited sample which may be derived from $\mathbf{x}_t + \gamma \mathbf{v}_i$.
Using the first-order Taylor expansion, $\tepsilont(\mathbf{x}'_t) \approx \tepsilont(\mathbf{x}_t) + \nabla_{\mathbf{x}_t} \tepsilont(\mathbf{x}_t) \cdot (\mathbf{x}'_t - \mathbf{x}_t)$,
we have an updated equation:
\begin{equation}
\label{eq:cpc}
\mathbf{x}'_t = \mathbf{x}_t + \frac{\sqrt{\alpha_t}}{1-\kappa \sqrt{1 - \alpha_t}} (\mathbf{x}'_0 - \mathbf{x}_0(\mathbf{x}_t)),
\end{equation}
where we use $\kappa = 0.99$. See \aref{appendixsec:editing} for a detailed derivation.

%% file: 4experiment.tex
\begin{figure}[!t]
    \centering
    \includegraphics[width=1\linewidth]{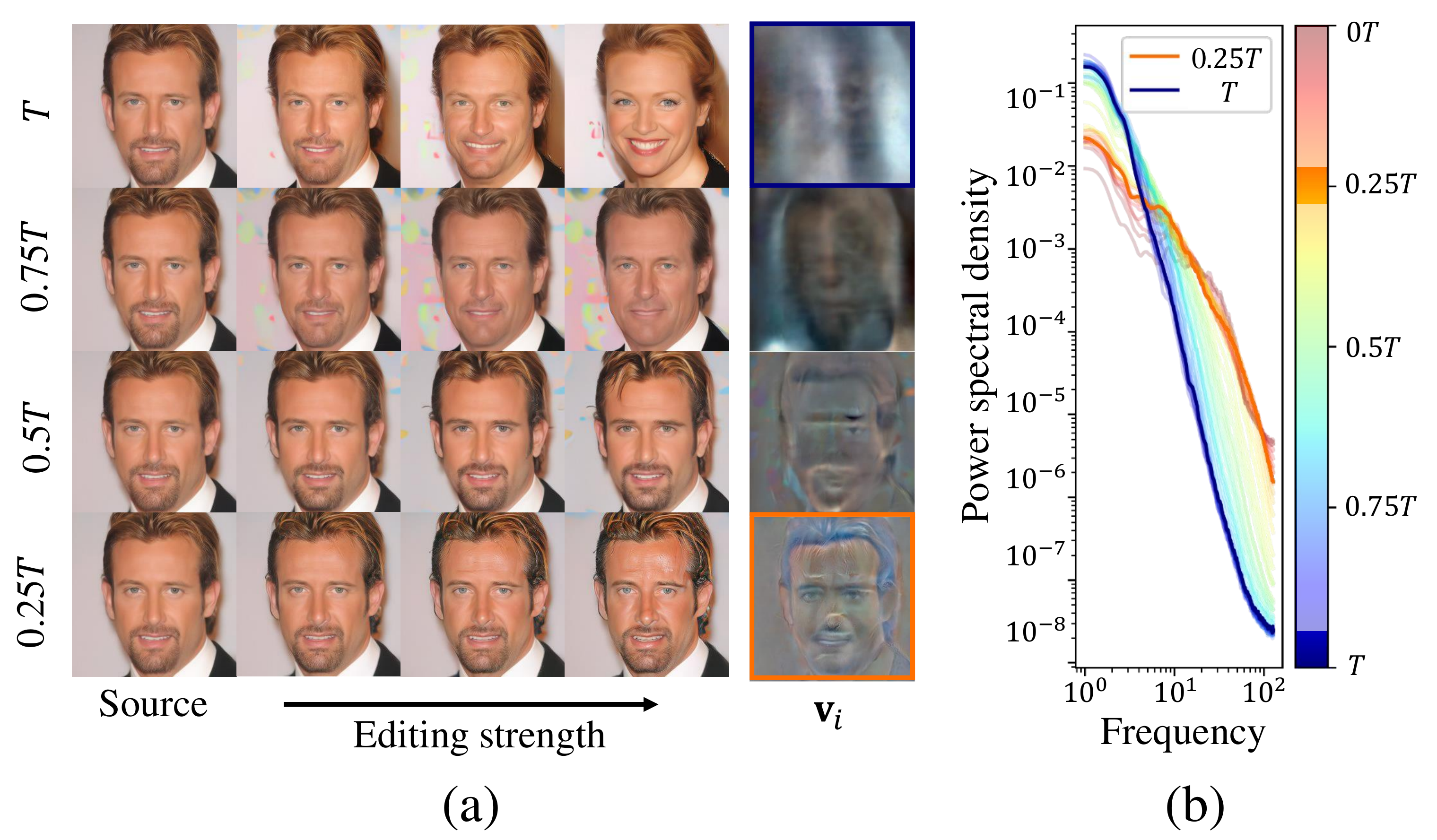}
    \vspace{-1em}
    \caption{
        \textbf{Comparison of the directions at different timesteps.} (a) Qualitative comparison showing that the directions in earlier timesteps edit coarse attributes while ones in later timesteps focus on high-frequency components.
        (b) Power spectral density (PSD) of $\mathbf{v}_i$. The PSD at $t=T$ (blue line) shows a larger portion of low-frequency signals, whereas the PSD at smaller $t$ (orange line) shows a larger portion of high-frequency signals.
    }
    \label{fig:PSD}
\end{figure}

\section{Experiments}

Thorough experiments demonstrate the usefulness of our method in various aspects.
The editing latent directions in \exspace{} found by our method include semantic changes and exhibit coarse-to-fine behavior (\sref{sec:local}). \exspace{} is a spherically curved space (\sref{sec:slerp}). Our method generalizes to stable diffusion (\sref{sec:stable}). Both the finding directions and the editing equation contribute to the nice properties of our method (\sref{sec:ablation}). Our method outperforms the existing methods (\sref{sec:comparison}).


\paragraph{Implementation details}
We validate our method and provide analyzes in CelebA-HQ \cite{karras2018progressive} for DDPM++ \cite{ho2020denoising, meng2021sdedit}, AFHQ-dog \cite{choi2018stargan} for iDDPM \cite{nichol2021improved}. All input images are from test sets in $256^2$ resolution. Quantitative results are from CelebA-HQ, unless otherwise noted. For Stable Diffusion ~\cite{rombach2022high}, we use ``Cyberpunk city" and ``Painting of Van Gogh" as text prompts to showcase the versatility of our method.
We use the official codes and pre-trained checkpoints for all baselines and keep the parameters \textit{frozen}. Further implementation details are deferred to \aref{supp:impl}.
The source code for our experiments is included in the supplementary materials, and will be publicly available upon publication.

\begin{figure}[!t]
    \centering
    \includegraphics[width=1.0\linewidth]{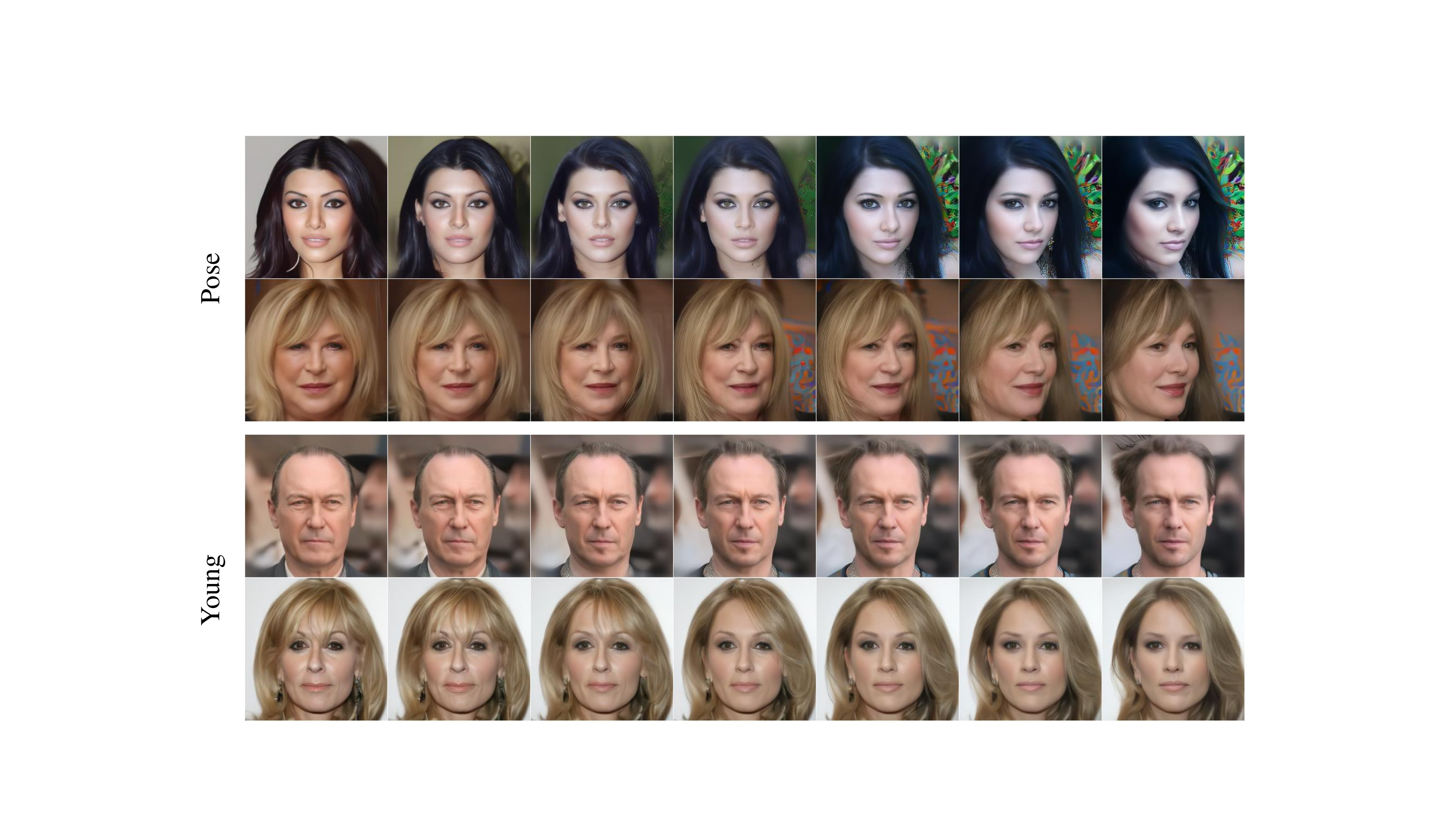}

    \centering
    \includegraphics[width=1.0\linewidth]{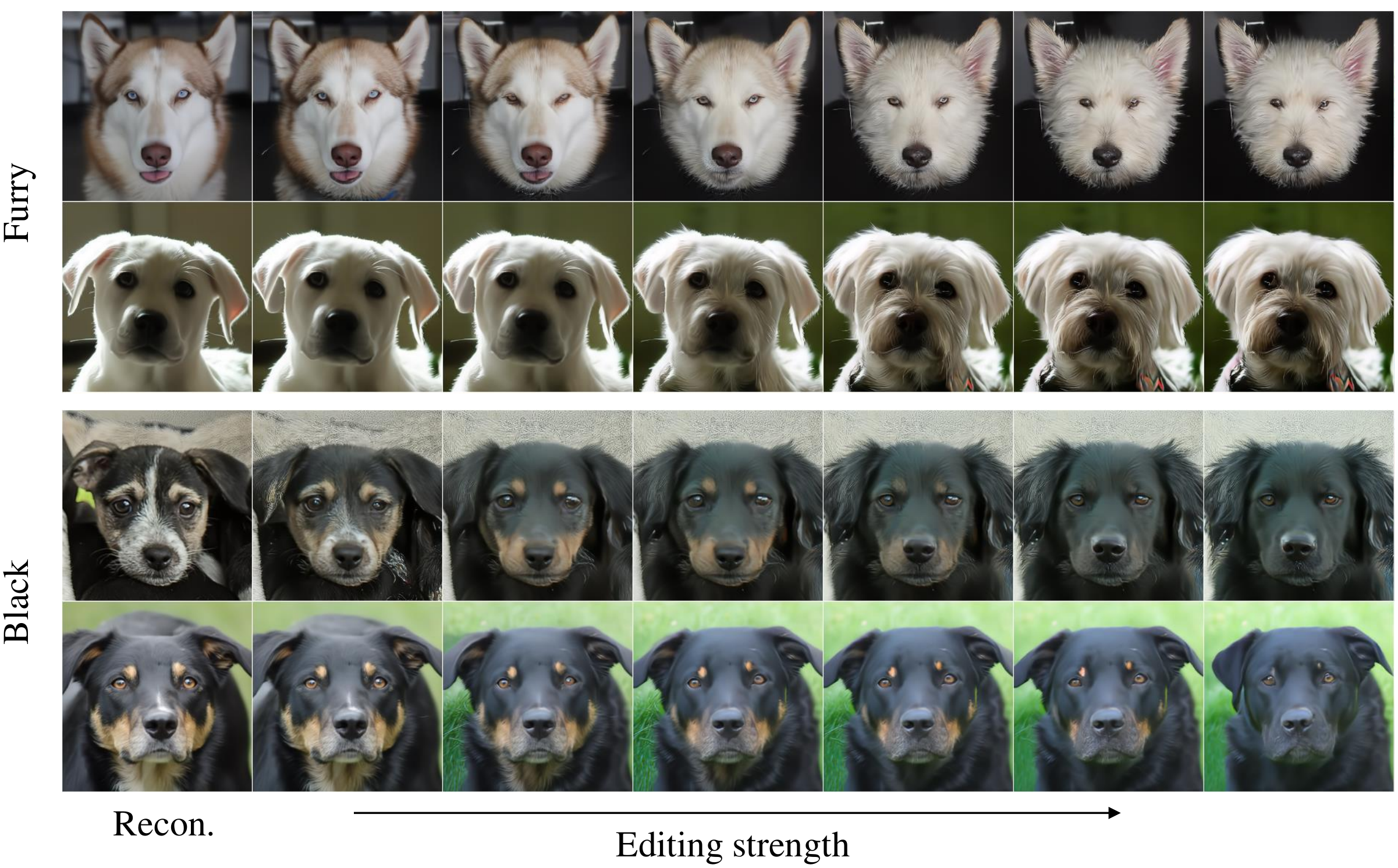}
    \vspace{-1em}
    \caption{
    \textbf{Example images edited with global semantic directions.} Consistent semantic changes in two rows validate the global semantic direction. The attributes are manually interpreted because the directions are not supervised.}
    \label{fig:global_basis_main}
\end{figure}

\subsection{Image manipulation}
\label{sec:local}

\paragraph{Semantic latent directions}
\fref{fig:local_basis} illustrates the example results edited by the directions found by our method \emph{without supervision} such as CLIP or a classifier. The directions clearly contain semantics such as gender, age, ethnicity, facial expression, breed, and texture. Interestingly, editing at timestep $T$ leads to coarse changes such as hair color, hair length, far breed. On the other hand, editing at the timestep $0.5T$ leads to fine changes such as make-up, hair texture, wrinkles, facial expression, and close breed. \aref{appendix:local} provides more examples.


\paragraph{Editing timing}
We further investigate the coarse-to-fine editing along the generative process from timestep $T$ to $0$. \fref{fig:PSD} (a) shows the example directions $\vv_i$ across different timesteps. At $T$, $\vv_i$ leads to coarse attribute changes in $\vx_0$ by blurry change in $\vx_T$. At $0.25T$, $\vv_i$ edits high-frequency details in both $\vx_0$ and $\vx_t$. \fref{fig:PSD} (b) shows the power spectral density (PSD) of $\vv_i$. We compute the PSD by taking $\vv_1, ..., \vv_{10}$ from 20 samples. The early timesteps contain a larger portion of low frequency than the later timesteps and the later timesteps contain a larger portion of high frequency. This phenomenon agrees with the tendency in the edited images. This results strengthens the common understanding of the timesteps~\cite{kwon2022diffusion,choi2022perception, daras2022multiresolution}.

\paragraph{Global semantic directions}
\fref{fig:global_basis_main} demonstrates that the global directions in $\vx_t$ lead to the same semantic changes, such as rotation, age, furriness, or color in different samples. It confirms that \exspace{} inherits the homogeneity of \ehspace{} via the pullback metric although \exspace{} is a metric-less space. \aref{appendix:global} provides more examples.


\begin{table}[t]
\caption{Semantic path length for lerp, slerp, and geodesic paths in CelebA-HQ for DDPM++.}
\label{tab:semantic_path_length}
\vskip 0.15in
\begin{center}
\begin{small}
\begin{tabular}{lc}
\toprule
Path & Semantic Path Length $(\mu \pm \sigma)$ \\
\midrule
lerp     & 10.29 $\pm$ 1.11 \\
slerp    & 7.69 $\pm$ 0.87 \\
geodesic & 5.98 $\pm$ 0.76 \\
\bottomrule
\end{tabular}
\end{small}
\end{center}
\vskip -0.1in
\end{table}

\begin{figure}[!t]
    \centering
    \includegraphics[width=1.0\linewidth]{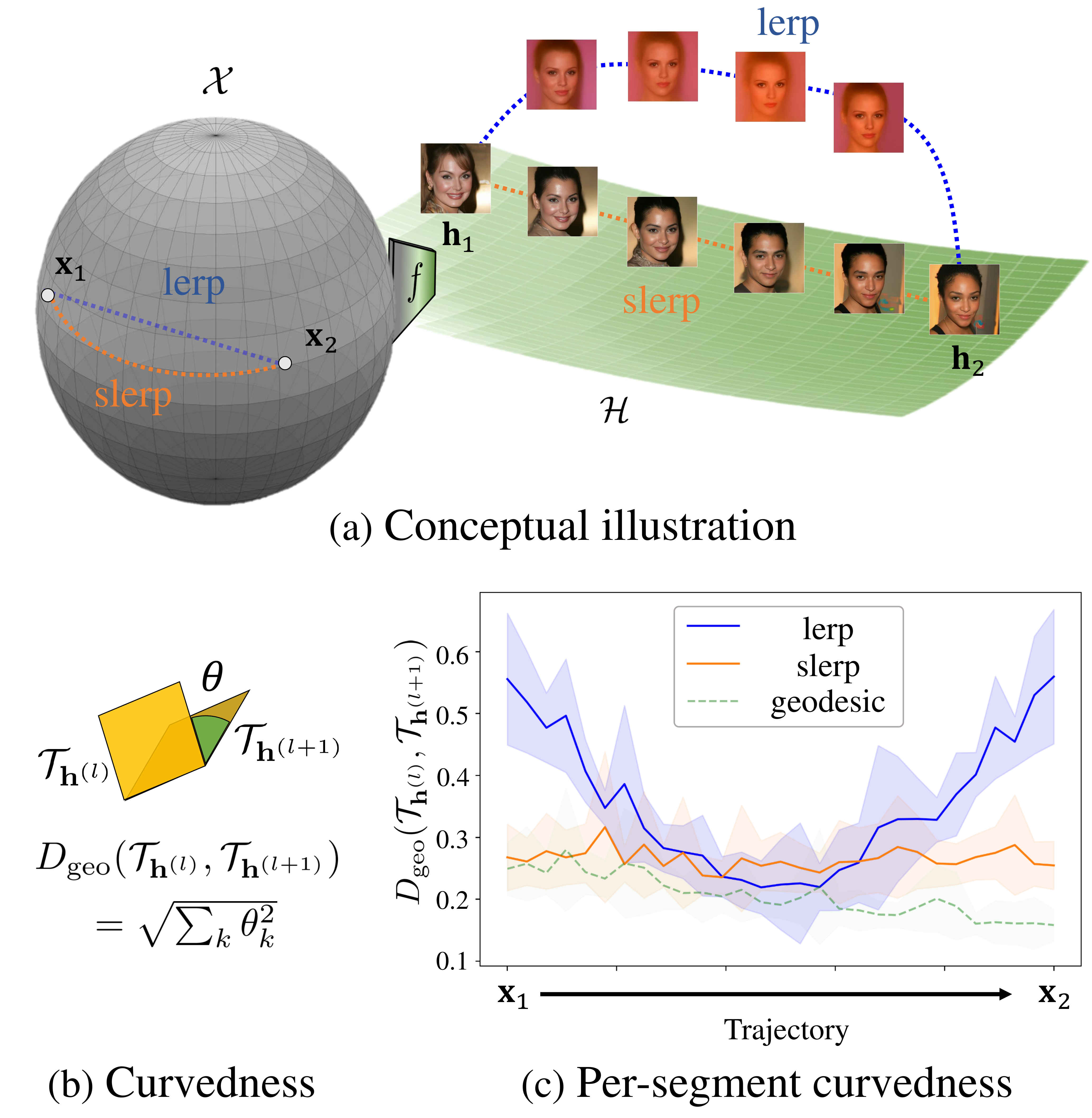}
    \vspace{-1em}
    \caption{
    \textbf{$\mathcal{X}$ is a curved manifold.}
    (a) Conceptual illustration of a linear path (lerp) and a spherical path (slerp) on the manifolds.
    (b) Curvedness of a line segment.
    (c) Per-segment distribution of curvedness along different paths. The lerp paths roughly have higher curvedness than slerp and reach similar curvedness to slerp. The slerp paths are closer to the geodesic shooting paths. It implies that $\mathcal{X}$ is a curved manifold. The shades depict $\pm$ 0.5 standard deviation. We use 50 segments for each path between $\vx_1$ and $\vx_2$.
    }
    \label{fig:slerp_lerp}
\end{figure}

\subsection{Curved manifold of DMs} 
\label{sec:slerp}
We present empirical grounds for the assumption in \sref{sec:method_local}: $\mathcal{X}$ is a curved manifold. Semantic path length between two points on a manifold is defined by the sum of the local warpage of the line segments which connects them along the manifold. We use \emph{geodesic metric}~\cite{choi2021not, ye2016schubert} to define the curvedness of a line segment $\{\vx^{(1)}, \vx^{(2)}\}$ as the angle between two tangent spaces centered at $\{\vh^{(1)}, \vh^{(2)}\}$: 
\begin{equation}
D_{\text{geo}}(\mathcal{T}_{\vh^{(1)}}, \mathcal{T}_{\vh^{(2)}}) = \sqrt{\sum_k \theta_k^2},
\end{equation}
where $\theta_k = \cos^{-1}(\sigma_k)$ denotes the $k$-th principle angle between $\mathcal{T}_{\vh^{(1)}}$ and $\mathcal{T}_{\vh^{(2)}}$.
The angle is visualized in \fref{fig:slerp_lerp} (b). Then, the semantic path length becomes $\sum_l D_\text{geo}(\mathcal{T}_{\vh^{(l)}}, \mathcal{T}_{\vh^{(l+1)}})$, where $l$ denotes the segment index in the path. We set the number of segments to $30$.
Then, the semantic path length increases as the path deviates further from the manifold.

\begin{figure}[!t]
    \centering
    \includegraphics[width=1\linewidth]{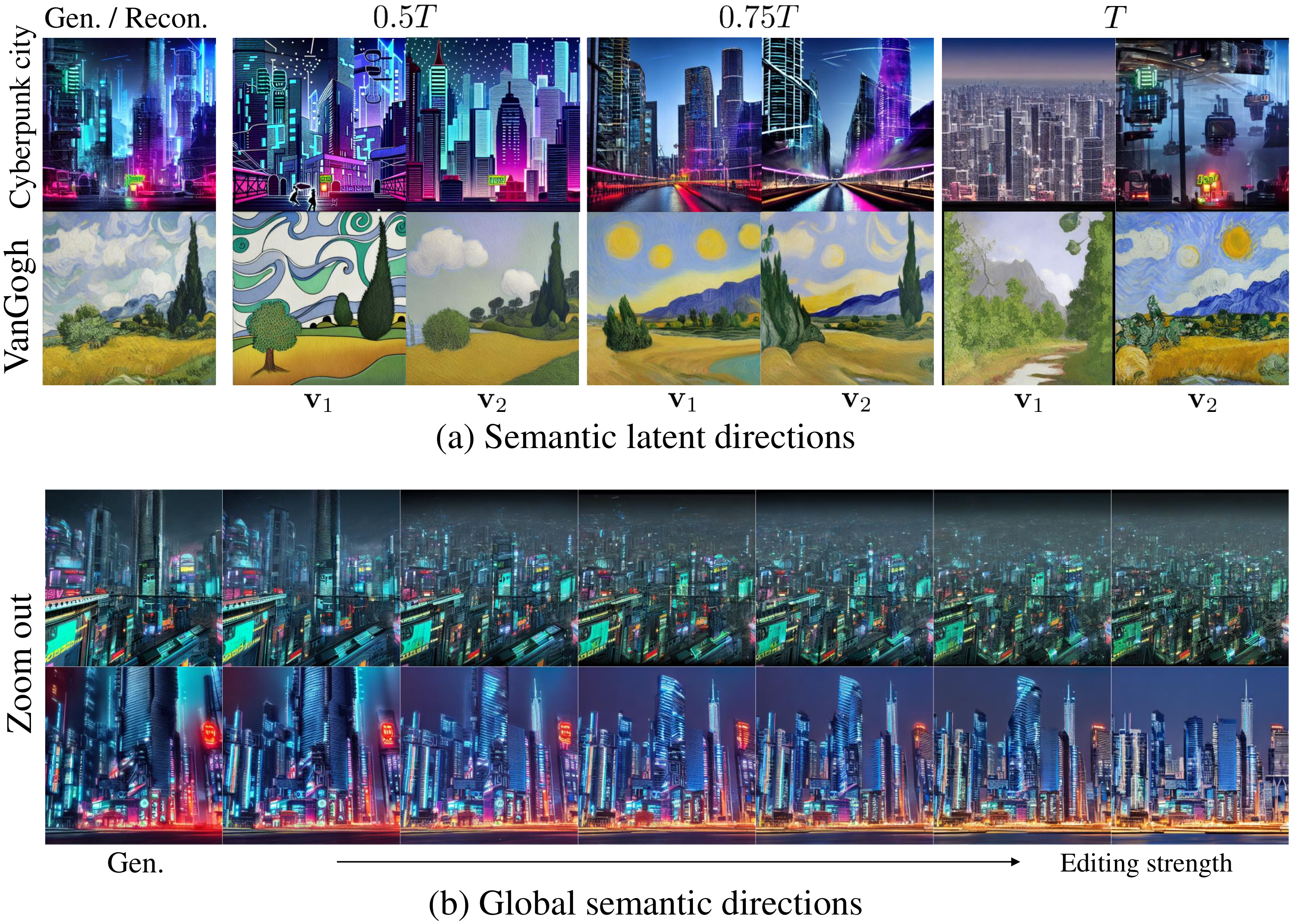}
    \vspace{-1em}
    \caption{
    \textbf{Generalization to Stable Diffusion.} 
    (a) Semantic latent directions successfully edit the images conditioned on ``Cyberpunk" and ``painting of Van Gogh". The leftmost images are the originals. Different directions and timesteps edit different attributes.
    (b) A global semantic direction consistently zooms out the different generated images.
    }
    \label{fig:ldm}
\end{figure}

To verify the assumption, we compare the semantic path lengths of different paths, e.g., linear path, spherical path, and geodesic shooting path. 
\fref{fig:slerp_lerp} (a) visualizes the manifold, linear path (lerp), and spherical path (slerp) and their corresponding path on $\mathcal{H}$ mapped by the function $f$. We computed the semantic path lengths for 50 randomly selected pairs of images. \tref{tab:semantic_path_length} shows that the semantic path length of slerp is smaller than lerp, indicating that the slerp path lies closer to the manifold than lerp, i.e., the manifold is curved. \fref{fig:slerp_lerp} (b) shows the distribution of the length of the segments along the path. Interestingly, the length of the lerp is high at the ends and shrinks to that of geodesics near the center. We suppose that the lerp path moves away from the original manifold and moves along another manifold.

Our semantic path length resembles the perceptual path length (PPL, \citet{karras2019style}) regarding the summation along the interpolation path.
PPL measures LPIPS \cite{zhang2018unreasonable} distance between resulting images along the path. Higher PPL between two latent variables indicates spikier interpolation of images accompanying artifacts. On the other hand, semantic path length measures how drastically the geometric structure changes between neighboring tangent spaces.

\subsection{Stable diffusion} 
\label{sec:stable}
This section demonstrates that our method is generalized to Stable Diffusion \cite{rombach2022high}. Our method extracts latent directions in the learned latent space $\vz_t$ using the same procedure. \fref{fig:ldm} (a) shows the edited images along different directions on various timesteps. The phenomena are similar to the image-based DMs: editing at $t=T$ provides coarse changes, and editing at later timesteps provides more fine texture-ish changes such as cartoonization.

Furthermore, \fref{fig:ldm} (b) shows that a global semantic direction leads to the same \textit{zoom-out} effect on different samples. Contrary to the global directions in the image-based DMs, the global directions are found within a text prompt, i.e., each text prompt has its own global directions. \aref{appendix:additional_results} provides more examples where we find some odd cases indicating that the learned latent space may not follow the same assumptions of the image-based DMs or the text guidance somehow twists the manifold.

\begin{figure}[!t]
    \centering
    \includegraphics[width=1\linewidth]{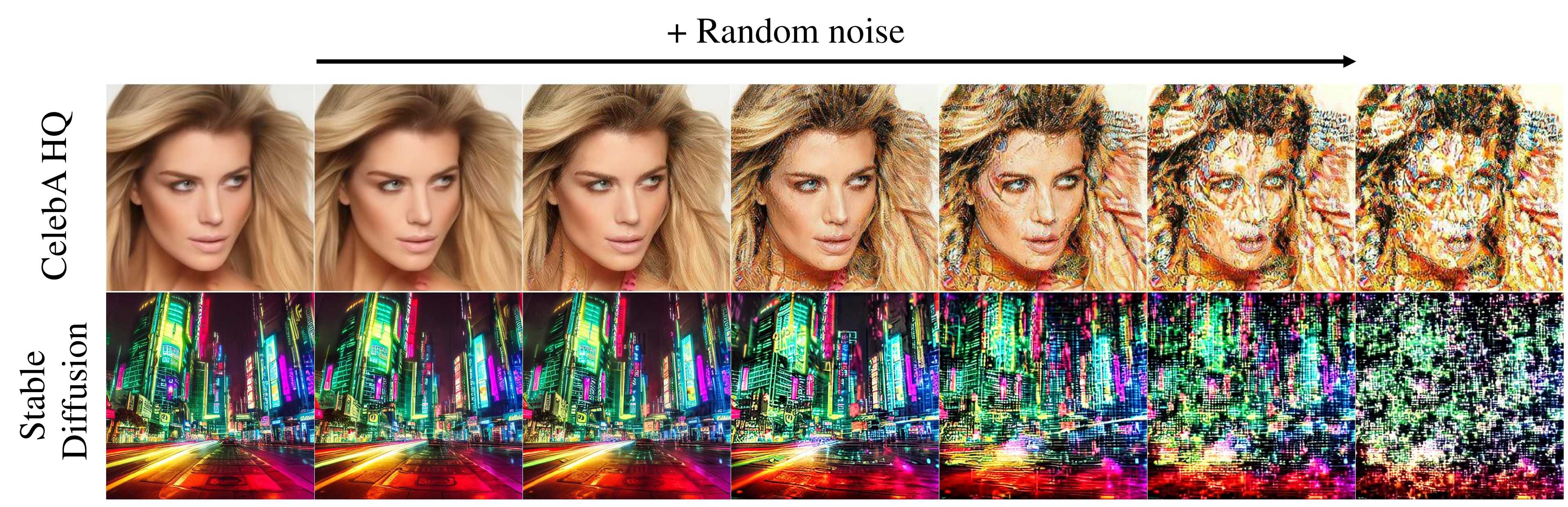}
    \vspace{-1em}
    \caption{
    \textbf{Importance of the discovered semantic directions.} Adding random directions instead of semantic directions severely distorts the resulting images. 
    }
    \label{fig:ablation_random_dx_ours}
\end{figure}

\begin{figure}[!t]
    \centering
    \includegraphics[width=1\linewidth]{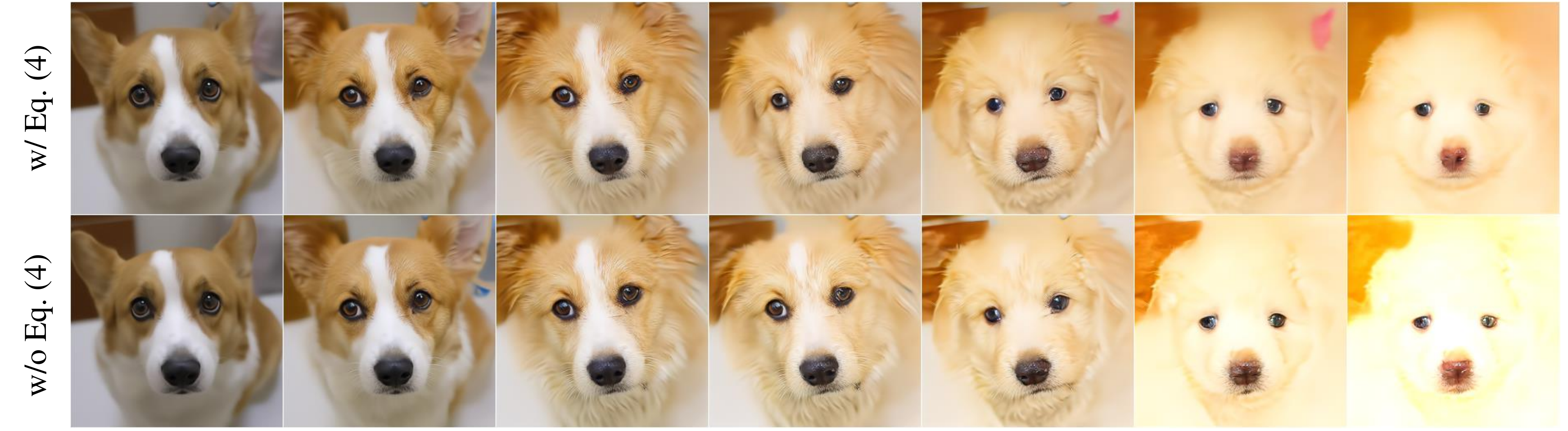}
    \vspace{-1em}
    \caption{
    \textbf{Importance of the normalization in \eref{eq:cpc}.} Removing the normalization leads to excessive saturation.
    }
    \label{fig:ablation_direct_dx_ours}
\end{figure}

\subsection{Ablation study}
\label{sec:ablation}
We provide ablation studies that include alternative approaches.
First, we edit images by applying random directions instead of semantic latent directions. \fref{fig:ablation_random_dx_ours} shows that random directions seriously degrade the images. 
This experiment validates the excellence of the latent directions found by our method.


\fref{fig:ablation_direct_dx_ours} demonstrates the necessity of normalization in \eref{eq:cpc}. While our full method produces plausible edited images even with extreme changes, removing the normalization leads to excessive saturation.

\begin{figure}[!t]
    \centering
    \includegraphics[width=1.\linewidth]{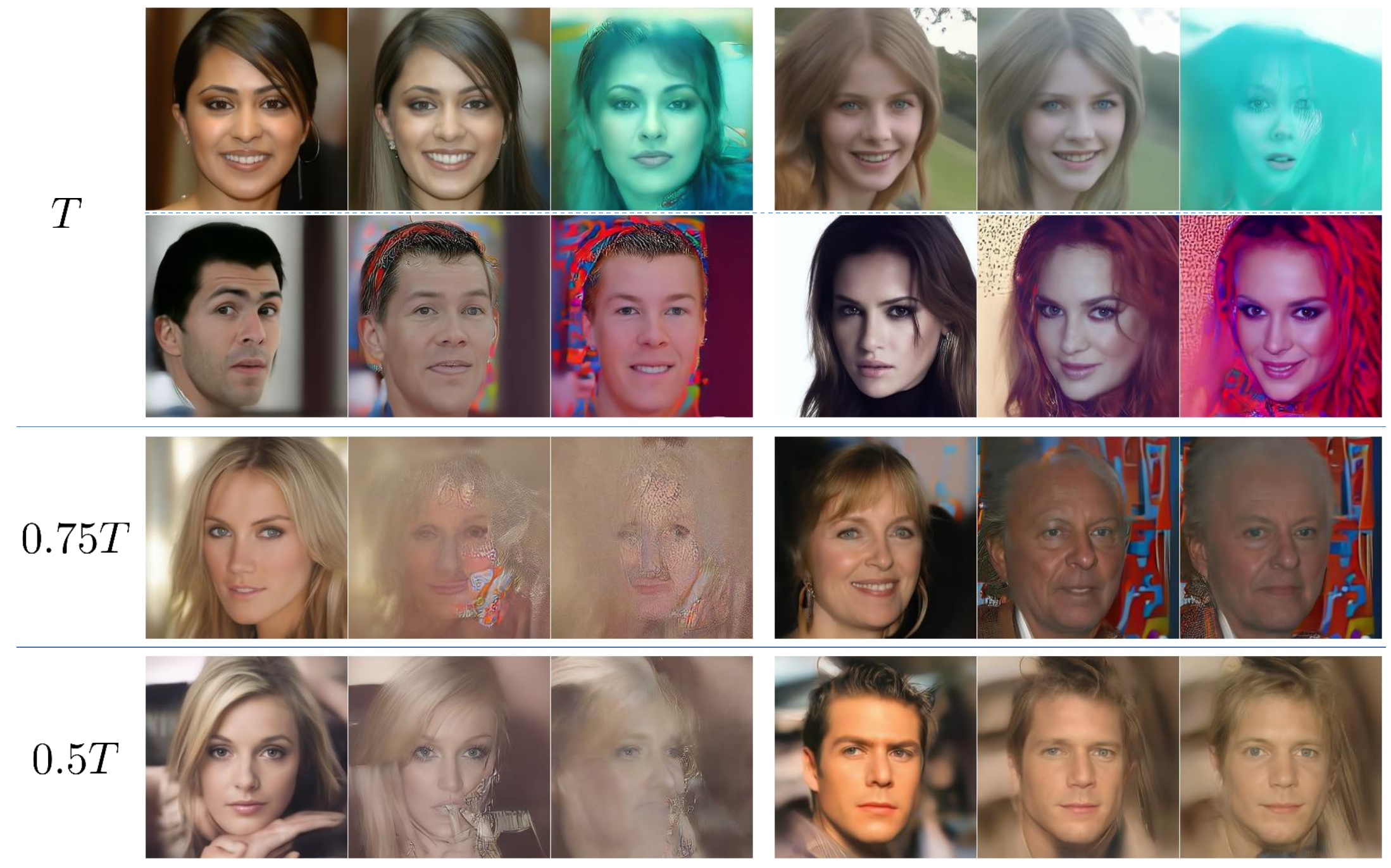}
    \vspace{-1em}
    \caption{
    \textbf{Inferiority of GANSpace on $\mathcal{H}$.} The GANSpace directions accompany severe distortion or entanglement while somewhat altering the attributes such as expression, rotation, and age.}
    \vspace{-1em}
    \label{fig:ganspace}
\end{figure}

\subsection{Comparison to other editing methods}
\label{sec:comparison}
As we introduce the first unsupervised editing in DMs, we compare our method with GANSpace \cite{harkonen2020ganspace} considering the mapping from $\mathcal{X}$ to $\mathcal{H}$ instead of $\mathcal{Z}$ to $\mathcal{W}$ in GANs. Accordingly, we find directions in $\mathcal{H}$ using PCA. \fref{fig:ganspace} shows their effects: they somewhat alter the attributes but accompany severe distortion or entanglement. On the contrary, our method finds the directions with the largest changes in $\mathcal{H}$ considering the geometrical structure leading to decent manipulation as shown in earlier results. \aref{appendixsec:comparison} describes more details for GANSpace.

%% file: 5conclusion.tex
\section{Discussion}
\begin{figure}[!t]
    \centering
    \includegraphics[width=1.0\linewidth]{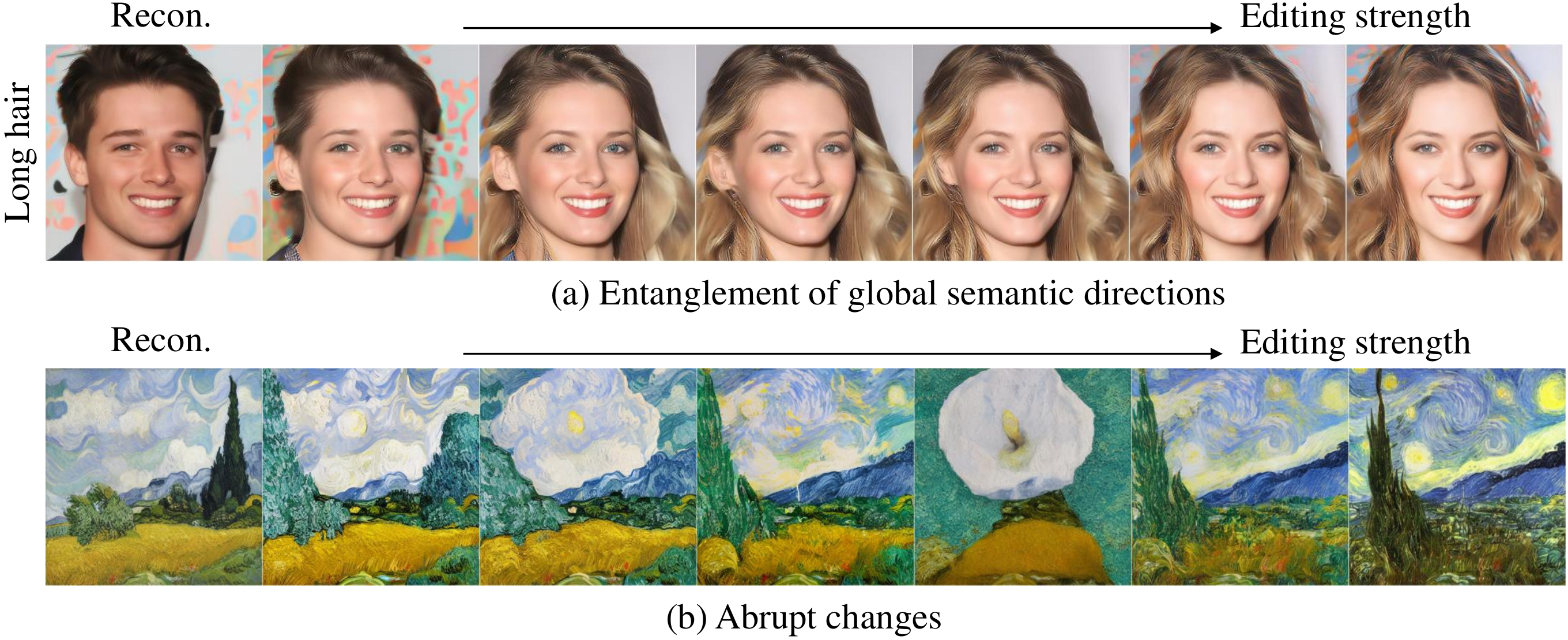}
    \vspace{-1em}
    \caption{
    \textbf{Limitations.} (a) Entanglement between attributes due to the dataset prior. (b) Abrupt changes in Stable Diffusion.}
    \label{fig:limitation}
\end{figure}

In this section, we provide additional intuitions and implications. It is interesting that our semantic latent directions usually convey disentangled attributes even though we do not adopt attribute annotation to enforce disentanglement. We suppose that decomposing the Jacobian of the encoder in the U-Nets naturally yields disentanglement to some extent. It grounds on the linearity of the intermediate feature space $\mathcal{H}$ in the U-Nets \cite{kwon2022diffusion}. However, it does not guarantee the perfect disentanglement and some directions are entangled. For example, the direction for long hair converts the male subject to female as shown in \fref{fig:limitation} (a). This kind of entanglement often occurs in other editing methods due to the dataset prior: there are few male faces with long hair.

Although we have shown that our method is also valid to Stable Diffusion, we still need more observation. It discovers less number of semantic latent directions and few directions occasionally convey abrupt changes during the editing procedure in Stable Diffusion as shown in \fref{fig:limitation} (b). We suppose that its learned latent space may have a more complex manifold than the image space \cite{arvanitidis2017latent}. Alternatively, the conditional DMs with classifier-free guidance or the cross-attention mechanism may add complexity on the manifold. Our future work includes analyzing the latent directions in the conditions such as text prompts or segmentation labels.

Despite these limitations, our method provides a significant advance in the field of image editing for DMs, and potential applications in a wide range of tasks.

\section{Conclusion}
In this work, we have proposed an unsupervised approach to extract semantic latent directions in $\mathcal{X}$, the latent space of diffusion models (DMs). 
Decomposing the Jacobian of the encoder in the U-Nets discovers the directions that manipulate mostly disentangled attributes. Our detailed analyses provide in-depth understanding of DMs: 1) different samples share the same latent directions in local tangent space leading to global semantic directions, 2) the generative process produces low-frequency components and adds high-frequency details, 3) the latent variable $\vx_t$ lives in a curved manifold, and 4) Stable Diffusion shares the similar intuitions with image-based DMs in the learned latent space.

Furthermore, we believe that better understanding the latent space of DMs will open up new possibilities for the development of DMs in useful applications, similar to how the arithmetic operations on the latent space of GANs has led to various follow-up research.

%% file: 6appendix.tex
\appendix
\onecolumn

\renewcommand{\thetable}{A\arabic{table}}
\renewcommand{\thefigure}{A\arabic{figure}}
\setcounter{figure}{0}
\setcounter{table}{0}

\section{Implementation Details}
\label{supp:impl}

\tref{tab:setting} summarizes various hyperparameter settings in our experiments. Specific details not covered in the main text are discussed in the following paragraphs.
\paragraph{Inversion step}
To obtain the latent code of a given image, we compute the latent code $\mathbf{x}_T$ using DDIM inversion. \cite{song2020denoising} The inversion step hyperparameter refers to the number of DDIM steps used to calculate the latent code. For stable-diffusion, we use classifier-free guidance. 
\paragraph{Low dimensional approximation ($n$)}
In our work, we employ a low-dimensional approximation of the tangent space. Rather than fixing the dimensionality at $n$, we determined to dynamically choose $n$ based on the distribution of eigenvalues. More specifically, we approximated the tangent space with dimensions corresponding to eigenvalues with cumulative density below a given threshold. As such, Table 1 presents the threshold rather than the dimensionality $n$. It worth note that, despite being determined dynamically, the actual values of $n$ has stable for various images. For example, for $t = T, 0.75T, 0.5T, 0.25T$, the values of $n$ were approximately 25, 50, 75, and 100, respectively.
\paragraph{Quality boosting ($t_{boost}$)}
While DDIM alone is capable of generating high-quality images, \citet{karras2022elucidating} showed that the inclusion of stochasticity improves image quality, and \citet{kwon2022diffusion} suggested the technique of adding stochasticity at the end of the generative process. We employ this technique in our experiments on CelebA-HQ and Stable-Diffusion after $t_{boost}$.
\paragraph{Stable-Diffusion}
In order to mitigate the influence of classifier-free guidance, the strength of the guidance, denoted as $w$, was set to zero, utilizing only the text-conditional model. \cite{ho2022classifier} When generating the original Cyberpunk city images, we set the guidance strength as $w = 7.5$. The prompts utilized for the Cyberpunk city images were ``Cyberpunk city" and for the Van Gogh paintings, the prompt used was ``painting of Van Gogh." Through the process of DDIM inversion, latent codes $\mathbf{x}_T$, were generated given the appropriate prompts for each image, with the guidance strength also set to zero (i.e., $\text{guidance scale} = 1$ in the code).

\section{Improved Editing Equation}
\label{appendixsec:editing}
\citet{song2020denoising} derived the following equation:
\begin{equation}
\label{eq:ddim}
\begin{aligned}
\sqrt{\alpha_t}\mathbf{x}_0 = \mathbf{x}_t - \sqrt{1-\alpha_t} \tepsilont(\mathbf{x}_t).
\end{aligned}
\end{equation}

Given inferred $\mathbf{x}_0(\mathbf{x}_t+\dx{}_i)$ under feature edition, our object is to find corrected ${\mathbf{x}'_t}$ that satisfies the above equation, $\sqrt{\alpha_t}{\mathbf{x}'_0} = \mathbf{x}'_t - \sqrt{1-\alpha_t} \tepsilont({\mathbf{x}'_t})$.
As stated in the main text, it is difficult to have exact ${\mathbf{x}'_t}$.
However, we can decompose the solution as ${\mathbf{x}'_t} = \mathbf{x}_t + d\mathbf{x}_t$, and then obtain the solution for the small $d\mathbf{x}_t$. This approximation follows as

\begin{align}
\sqrt{\alpha_t}{\mathbf{x}'_0}  &= \mathbf{x}'_t - \sqrt{1-\alpha_t}
\tepsilont({\mathbf{x}'_t}) \\
\Rightarrow 
\sqrt{\alpha_t}(\mathbf{x}_0(\mathbf{x}_t) + d\mathbf{x}_0) &= \mathbf{x}_t + d\mathbf{x}_t - \sqrt{1-\alpha_t} \tepsilont(\mathbf{x}_t + d\mathbf{x}_t) \\
\Rightarrow 
\sqrt{\alpha_t}(\mathbf{x}_0(\mathbf{x}_t) + d\mathbf{x}_0) &= \mathbf{x}_t + d\mathbf{x}_t - \sqrt{1-\alpha_t} \big(\tepsilont(\mathbf{x}_t) + \nabla_{\mathbf{x}_t} \tepsilont \cdot d\mathbf{x}_t \big)  \\
\Rightarrow 
\sqrt{\alpha_t}d\mathbf{x}_0 &= d\mathbf{x}_t - \sqrt{1-\alpha_t} \nabla_{\mathbf{x}_t} \tepsilont \cdot d\mathbf{x}_t  \\
\Rightarrow 
\sqrt{\alpha_t}d\mathbf{x}_0 &= d\mathbf{x}_t - \kappa \sqrt{1-\alpha_t} d\mathbf{x}_t  \\
\therefore
d\mathbf{x}_t &= \frac{\sqrt{\alpha_t}}{1 - \kappa \sqrt{1-\alpha_t}} d\mathbf{x}_0
\end{align}

\begin{figure}[!t]
    \centering
    \includegraphics[width=0.6\linewidth]{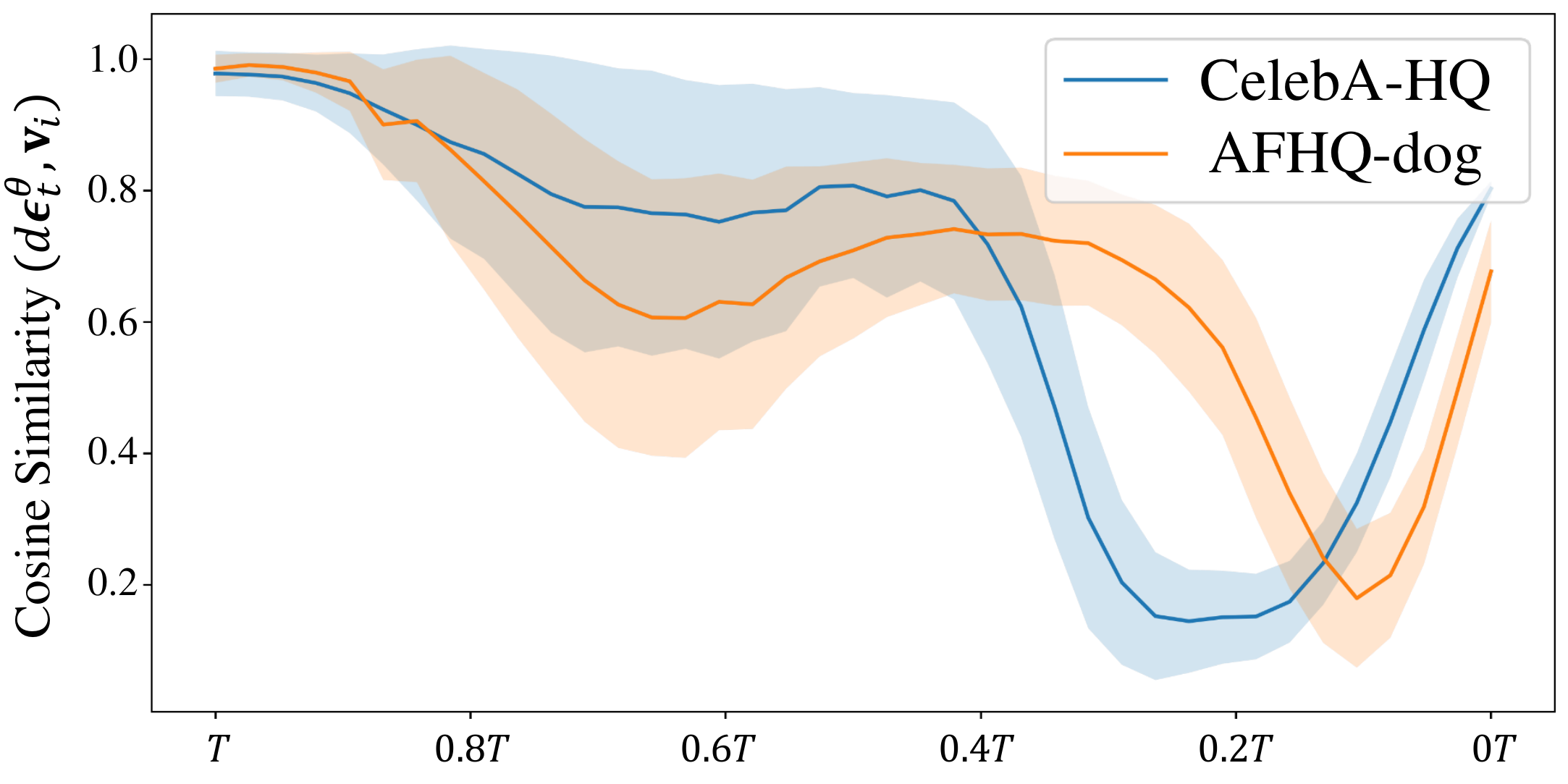}
    \caption{
        Cosine similiarity between $(d\tepsilont, \mathbf{v}_i)$. The shaded region represents the mean $\pm$ standard deviation of the measurements. The results depicted in the figure were obtained by measuring 100 samples from $\mathbf{x} \sim \mathcal{N}(0, \mathrm{I})$.
    }
    \label{fig:jac_w_Id}
\end{figure}

In the third line of the derivation, we used a first-order Taylor expansion of $\tepsilont$. In the fourth line, we eliminated dominant terms on both sides using Eq.(6). In the fifth line, we approximated the Jacobian matrix as an identity matrix, $\nabla_{\mathbf{x}_t} \tepsilont \approx \kappa \mathrm{I}$.
This approximation enables us to obtain $\mathbf{x}'_t$.
it is important to note that the Jacobian is multiplied by $\sqrt{1-\alpha_t}$.
Therefore, the component works only for $t$ close to $T$, because otherwise $\sqrt{1-\alpha_t}$ vanishes.
Then, it is sufficient to show that our approximation works well in the range of $t$ close to $T$.
We examined the validity of our approximation numerically through the following equation,
\begin{equation}
\nabla_{\mathbf{x}_t} \tepsilont \approx \frac{\tepsilont(\mathbf{x}_t + d\mathbf{x}_t) - \tepsilont(\mathbf{x}_t)}{||d\mathbf{x}_t||}
\end{equation}
where $d\mathbf{x}_t = \dx{}_i$.
Then, the approximation of $\nabla_{\mathbf{x}_t} \tepsilont \approx \kappa \mathrm{I}$ represents a good alignment between two vectors of $d\tepsilont$ and $\dx{}_i$. 
We confirmed the good alignment using their cosine similarity (\fref{fig:jac_w_Id}). We use $\kappa = 0.99$, since it is a representative value in the range $t \approx T$. This value also serves to prevent the vector $d\mathbf{x}$ from becoming excessively small, when $\sqrt{\alpha_t} \approx 0$.




\begin{table}[t]
\caption{Hyper-parameter settings.}
\label{tab:setting}
\vskip 0.1in
\begin{center}
\begin{small}
\begin{tabular}{lcccccc}
\toprule
Experiment & $t_{edit}$ & $\gamma$ & inversion step & threshold ($n$) & $t_{boost}$ & guidance strength ($w$) \\
\midrule
CelebA-HQ           & $T$       & 0.0025 & 40 & 0.5 & 0.15$T$  & $\times$\\
                    & $0.75T$   & 0.0125 & 40 & 0.5 & 0.15$T$  & $\times$\\
                    & $0.5T$    & 0.2500 & 40 & 0.5 & 0.15$T$  & $\times$\\
                    & $0.25T$   & 2.5000 & 40 & 0.5 & 0.15$T$  & $\times$\\
AFHQ-dog            & $T$       & 0.0025 & 80 & 0.5 & $\times$ & $\times$ \\
                    & $0.75T$   & 0.0100 & 80 & 0.5 & $\times$ & $\times$ \\
                    & $0.5T$    & 0.2500 & 80 & 0.5 & $\times$ & $\times$ \\
                    & $0.25T$   & 2.5000 & 80 & 0.5 & $\times$ & $\times$ \\
Stable-Diffusion    & $T$       & 0.025 & 80 & 0.25 & 0.15$T$  & 0 \\
                    & $0.75T$   & 0.100 & 80 & 0.25 & 0.15$T$  & 0 \\
                    & $0.5T$    & 0.500 & 80 & 0.25 & 0.15$T$  & 0 \\
                    & $0.25T$   & 2.5000 & 80 & 0.25 & 0.15$T$ & 0  \\
\bottomrule
\end{tabular}
\end{small}
\end{center}
\vskip -0.1in
\end{table}

\section{Comparison Details}
\label{appendixsec:comparison}

Since unsupervised editing is not available for DMs, we consider GANSpace for image editing. 
The spaces of $\mathcal{Z}$ and $\mathcal{W}$ of GAN correspond to $\mathcal{X}$ and $\mathcal{H}$ of DM, respectively. We use 1k random images with DDIM generative process for GANSpace.
Note that the GANSpace method is obtaining directions in $\mathcal{W}$ thus we used GANSpace to add directions directly to \ehspace{}. In addition to what \sref{sec:comparison} provides, the editing direction, extracted by the GANSpace, primarily alters colors in images. This suggests that simply collecting every $\mathbf{h}$ in $\mathcal{H}$ and extracting their principal axes may find poor feature directions that may control just overall color.


\clearpage
\section{Algorithms}
\label{appendixsec:algorithm}

\begin{algorithm}[ht!]
\caption{Feature Direction}
\label{alg:local_basis}
\begin{algorithmic}[1]
\REQUIRE {latent variable $\mathbf{x}$, timestep $t$, U-Net encoder $f : \mathcal{X} \times T \rightarrow \mathcal{H}$, Feature direction index $i$}
\STATE {$J$             = Jacobian($f(\cdot, t)$)($\mathbf{x}$)}
\STATE $U, S, V^{\tran}$ = SingularValueDecomposition($J$) 
\STATE {$\mathbf{v}_i, \mathbf{u}_i$ = $V^{\tran}$[$i$, :], $U$[:, $i$]}
\STATE {{\bfseries Return} $\mathbf{v}_i, \mathbf{u}_i$}
\end{algorithmic}
\end{algorithm}

\begin{algorithm}[ht!]
\caption{Global Feature Direction}
\label{alg:global_basis}
\begin{algorithmic}[1]
\REQUIRE {latent variable $\mathbf{x}$, U-Net encoder $f : \mathcal{X} \times T \rightarrow \mathcal{H}$, low-dimensional approximation $n$, number of local bases $L$, Global feature direction $i$}
\FOR{$l=1$ {\bfseries to} $L$}
\STATE $\mathbf{x} \sim \mathcal{N}(0, \mathrm{I})$
\STATE {$J$             = Jacobian($f(\cdot, T)$)($\mathbf{x}$)}
\STATE $U^{(l)}$, $\cdot$, $\cdot$  = SingularValueDecomposition($J$) 
\STATE $U^{(l)}$ = $U^{(l)}$[:, :$n$]
\ENDFOR
\FOR{$l=1$ {\bfseries to} $L$}
    \IF {$l=1$}
        \STATE $\bar{U}$ = $U^{(l)}$
    \ELSE
        \FOR{ $m=1$ {\bfseries to} $n$}
            \STATE $U^{(l)}$ = SortBySimilarity($U^{(l)}$, $\bar{U}$)
            \STATE $\bar{U}$ = $\bar{U}$ + $\sum_k \text{sign}(\sum_{c} U^{(l)}_{ck}\bar{U}_{ck}) U^{(l)}_{\cdot k}$
        \ENDFOR
    \ENDIF
\ENDFOR
\STATE $\bar{U}$ = $\frac{1}{L} \bar{U}$
\STATE $\bar{\mathbf{u}_i}$ = $\bar{U}$[:, $i$]
\STATE {{\bfseries Return} $\bar{\mathbf{u}}_i$}
\end{algorithmic}
\end{algorithm}

\begin{algorithm}[ht!]
\caption{Tangent Space Projection}
\label{alg:local_projection}
\begin{algorithmic}[1]
\REQUIRE latent variable $\mathbf{x}$, timestep $t$, U-Net encoder $f : \mathcal{X} \times T \rightarrow \mathcal{H}$, \ehspace{} direction $\mathbf{u}_i$, low-dimensional approximation $n$. 
\STATE $J_{\mathbf{x}}$ = Jacobian($f(\cdot, t)$)($\mathbf{x}$)
\STATE $U, S, V^{\tran}$  = SingularValueDecomposition($J_\mathbf{x}$) 
\STATE $U, V^{\tran}$     = $U$[:, :$n$], $V^{\tran}$[:$n$, :]
\STATE $\mathbf{u}'_i$  = $UU^{\tran}\mathbf{u}_i$
\STATE $\mathbf{u}'_i$ /= $||\mathbf{u}'_i||$
\STATE $\mathbf{v}'_i$  = $VU^{\tran} \mathbf{u}'_i$
\STATE {\bfseries Return} $\mathbf{v}'_i, \mathbf{u}'_i$
\end{algorithmic}
\end{algorithm}

\begin{algorithm}[ht!]
\caption{Eq. (4)}
\label{alg:cpc}
\begin{algorithmic}[1]
\REQUIRE latent variable $\mathbf{x}$, timestep $t$, map from $\mathbf{x}_t$ to predicted $\mathbf{x}_0$ $P : \mathcal{X}_t \rightarrow \mathcal{X}_0$ edit step size $\gamma$, feature direction $\mathbf{v}_i$
\STATE $\mathbf{x}_0', \mathbf{x}_0 = P(\mathbf{x}_0 + \gamma \mathbf{v}_i), P(\mathbf{x}_0)$
\STATE $\mu_{\mathbf{x}_0'}, \mu_{{\mathbf{x}}_0}$ = Mean($\mathbf{x}_0'$), Mean($\mathbf{x}_0$)
\STATE $\mathbf{x}_0'$ = $\mu_{\mathbf{x}_0'} + (\mathbf{x}_0' - \mu_{\mathbf{x}_0'}) \frac{\text{StandardDeviation}({\mathbf{x}}_0 - \mu_{{\mathbf{x}}_0})}{\text{StandardDeviation}(\mathbf{x}_0' - \mu_{{\mathbf{x}}_0'})}$
\STATE $d\mathbf{x}$ = $\frac{\sqrt{\alpha_t}}{1 - 0.99 \sqrt{1-\alpha_t}}$ $(\mathbf{x}_0'- \mathbf{x}_0)$
\STATE {\bfseries Return} $d\mathbf{x}$
\end{algorithmic}
\end{algorithm}

\begin{algorithm}[ht!]
\caption{Total Editing Process}
\label{alg:total_alg}
\begin{algorithmic}[1]
\REQUIRE latent variable $\mathbf{x}$, timestep $t$, U-Net encoder $f : \mathcal{X} \times T \rightarrow \mathcal{H}$, (Global) feature direction index $i$, low-dimensional approximation $n$, edit step size $\gamma$, edit iteration $N_{\text{iter}}$
\IF {Use Global Feature Direction}
    \STATE $\mathbf{u}_i$ = GlobalFeatureDirection($\mathbf{x}$, $f$, $n$, $L$, $i$)
\ELSE
    \STATE $\mathbf{u}_i$ = FeatureDirection($\mathbf{x}$, $t$, $f$, $i$)
\ENDIF
\FOR {$edit=1$ {\bfseries to} $N_{\text{iter}}$}
    \STATE $\mathbf{v}_i, \mathbf{u}_i$ = TangentSpaceProjection($\mathbf{x}$, $t$, $f$, $\mathbf{u}_i$, $n$)
    \STATE $d\mathbf{x}$ = Eq. (4)($\mathbf{x}$, $t$, $P$, $\gamma$, $\mathbf{v}_i$)
    \STATE $\mathbf{x}$ = $\mathbf{x}$ + $d\mathbf{x}$
\ENDFOR
\STATE {\bfseries Return} $\mathbf{x}$
\end{algorithmic}
\end{algorithm}

\clearpage

\section{Additional results}
\label{appendix:additional_results}
\subsection{Feature direction}
\label{appendix:local}

\begin{figure}[!h]
    \centering
    \includegraphics[width=1.0\linewidth]{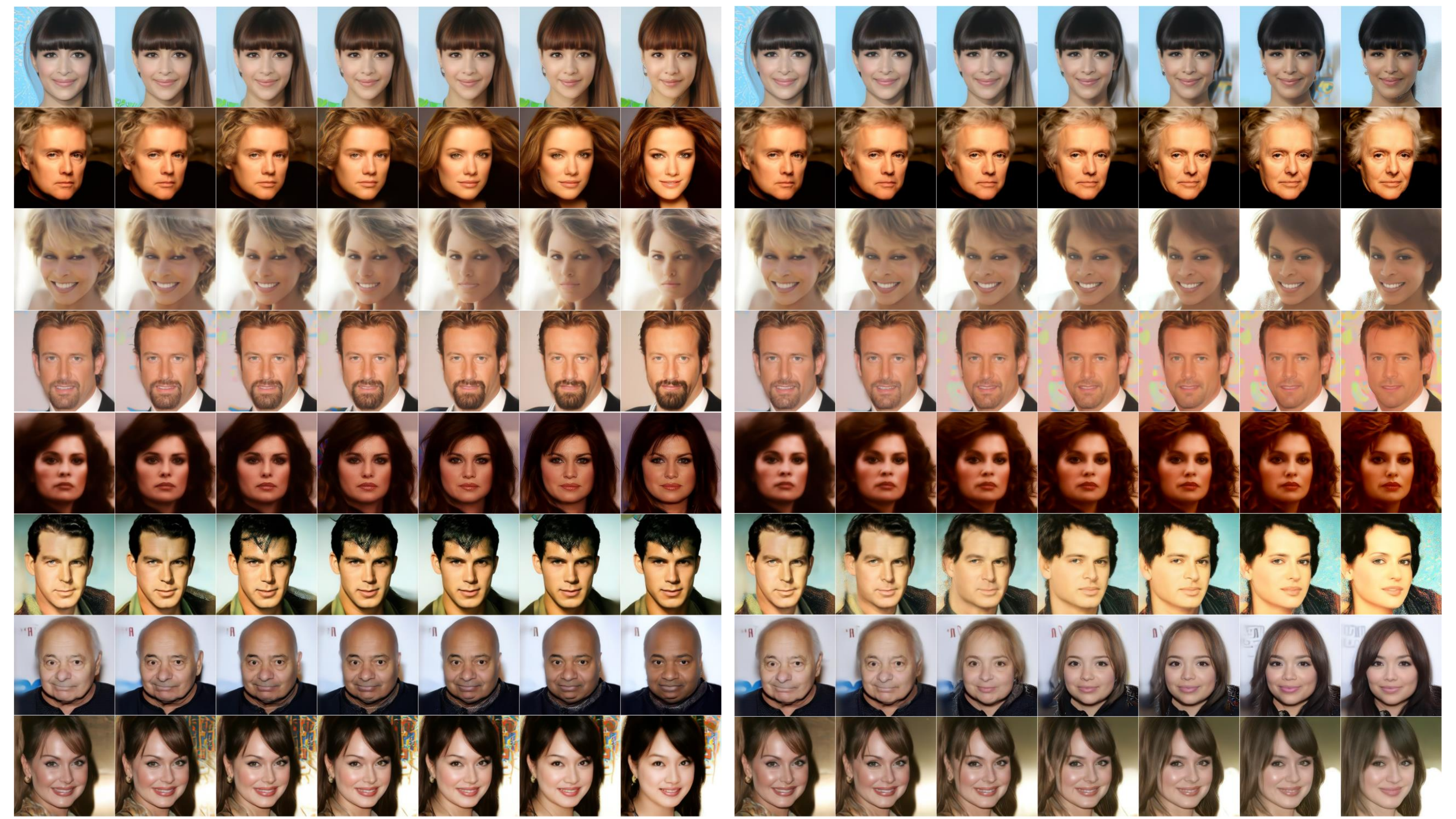}
    \caption{
    \textbf{$\boldsymbol{t=T}$.} A selection of interpretable edits discovered by our feature direction in CelebA-HQ. The image on the far left represents the reconstructed original image, while the subsequent images demonstrate the interpretable edits that have been made to it.}
    \label{fig:celeba_local_appendix}
\end{figure}

\begin{figure}[!t]
    \centering
    \includegraphics[width=1.0\linewidth]{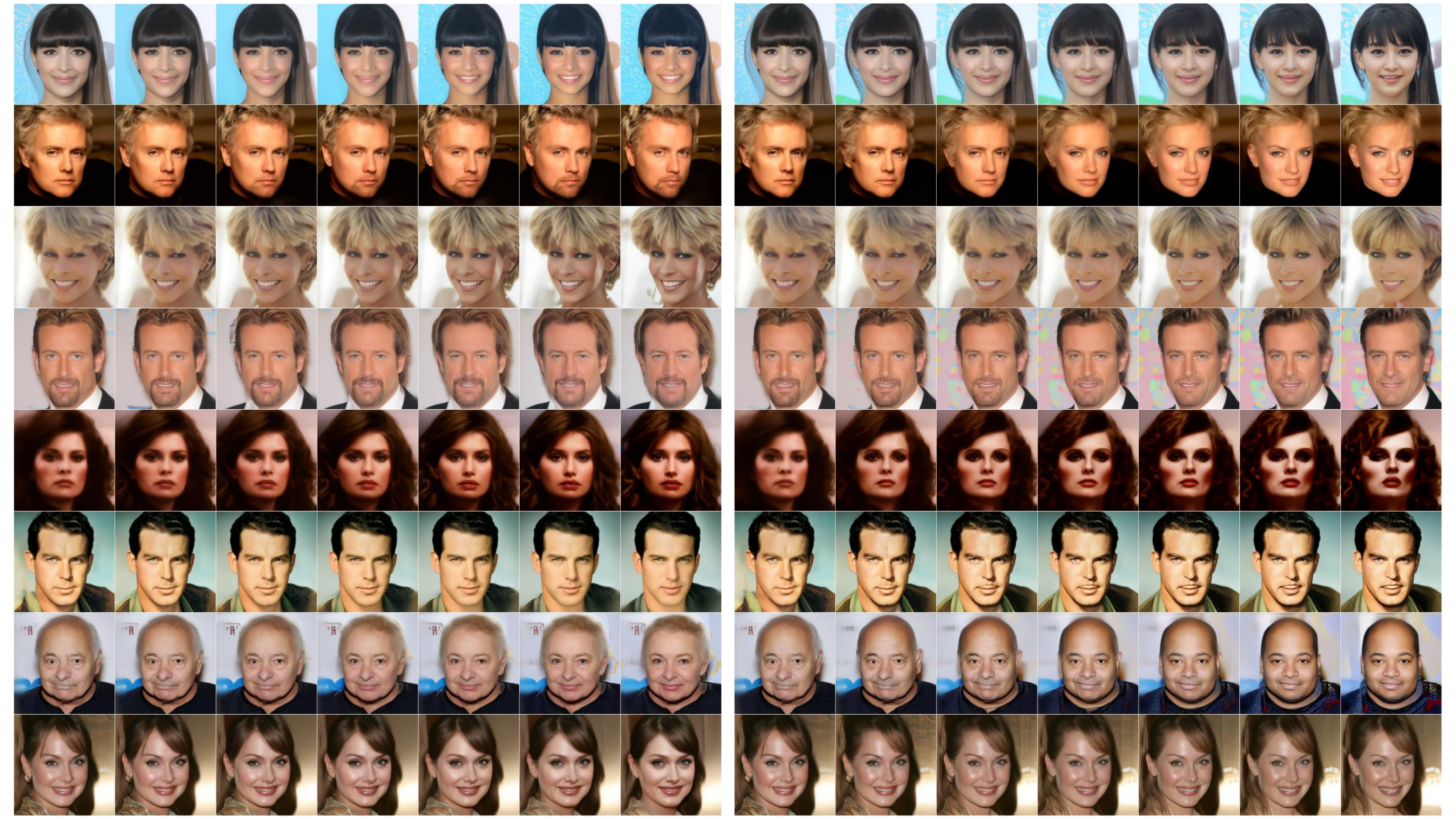}
    \caption{
    \textbf{$\boldsymbol{t=0.75T}$.} A selection of interpretable edits discovered by our feature direction in CelebA-HQ. The image on the far left represents the reconstructed original image, while the subsequent images demonstrate the interpretable edits that have been made to it.}
    \label{fig:celeba_local_appendix_75}
\end{figure}

\begin{figure}[!t]
    \centering
    \includegraphics[width=1.0\linewidth]{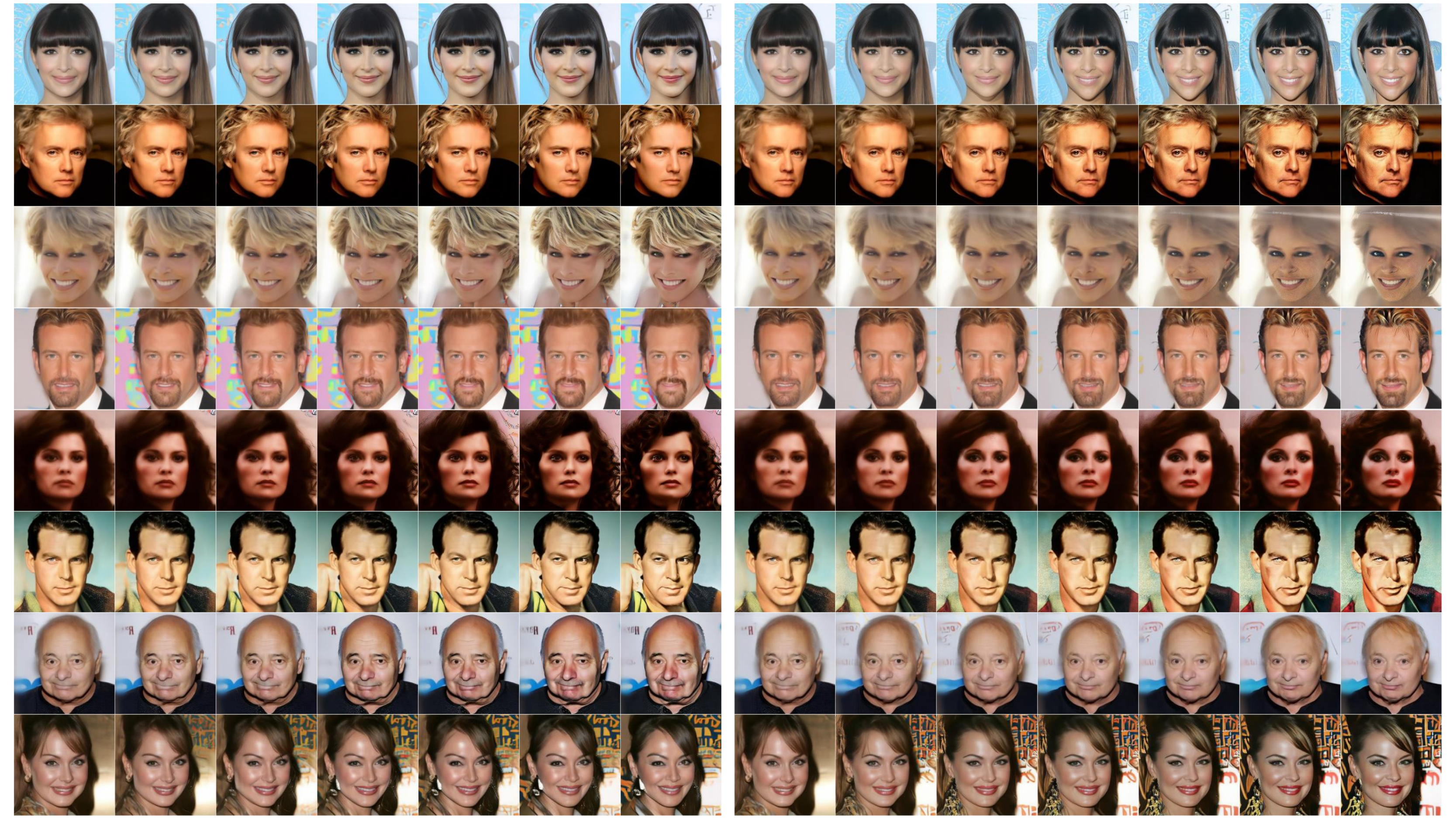}
    \caption{
    \textbf{$\boldsymbol{t=0.5T}$.} A selection of interpretable edits discovered by our feature direction in CelebA-HQ. The image on the far left represents the reconstructed original image, while the subsequent images demonstrate the interpretable edits that have been made to it.}
    \label{fig:celeba_local_appendix_50}
\end{figure}

\begin{figure}[!h]
    \centering
    \includegraphics[width=1.0\linewidth]{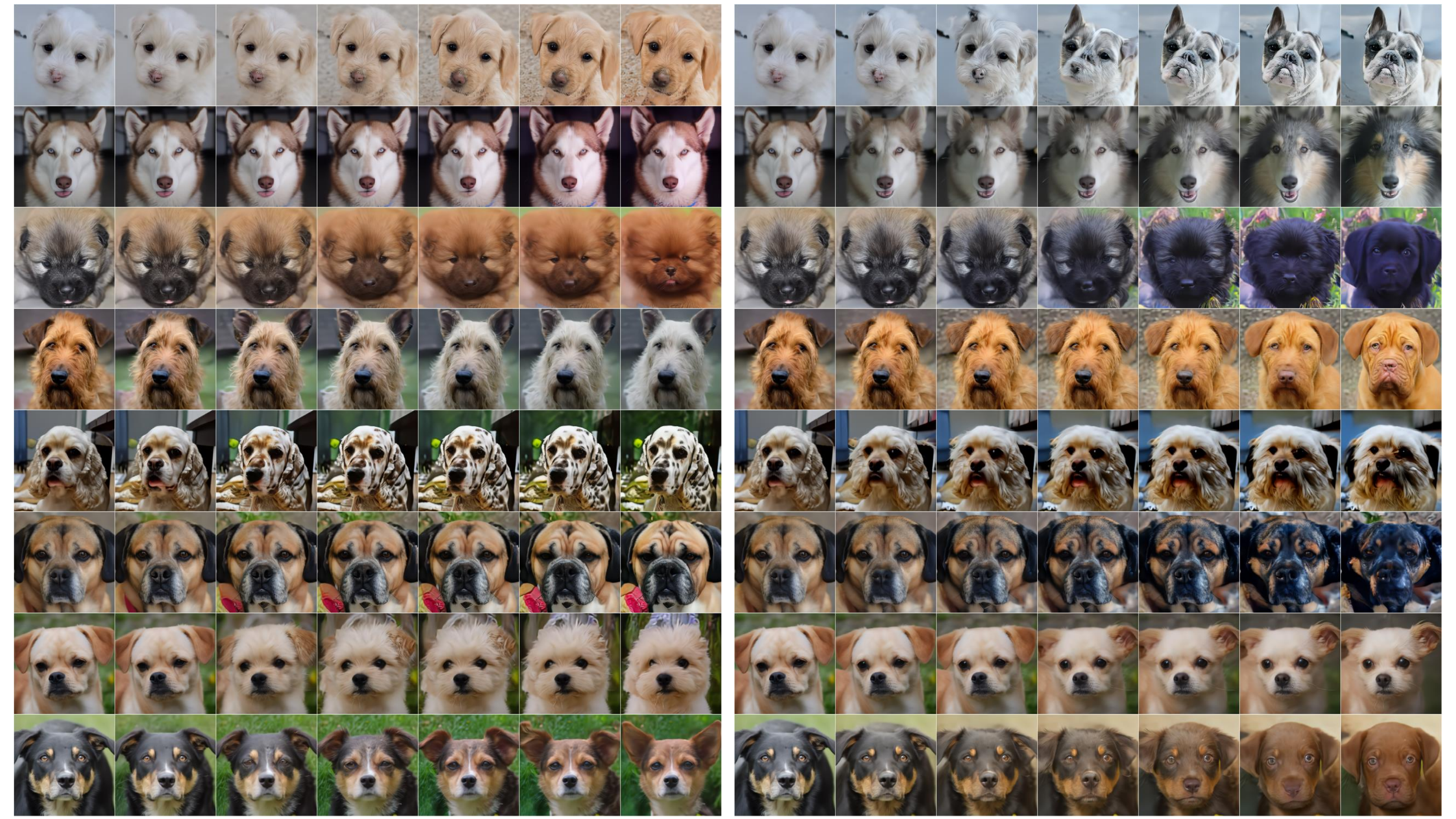}
    \caption{
    \textbf{$\boldsymbol{t=T}$.} A selection of interpretable edits discovered by our feature direction in AFHQ. The image on the far left represents the reconstructed original image, while the subsequent images demonstrate the interpretable edits that have been made to it.}
    \label{fig:afhq_local_appendix}
\end{figure}

\begin{figure}[!h]
    \centering
    \includegraphics[width=1.0\linewidth]{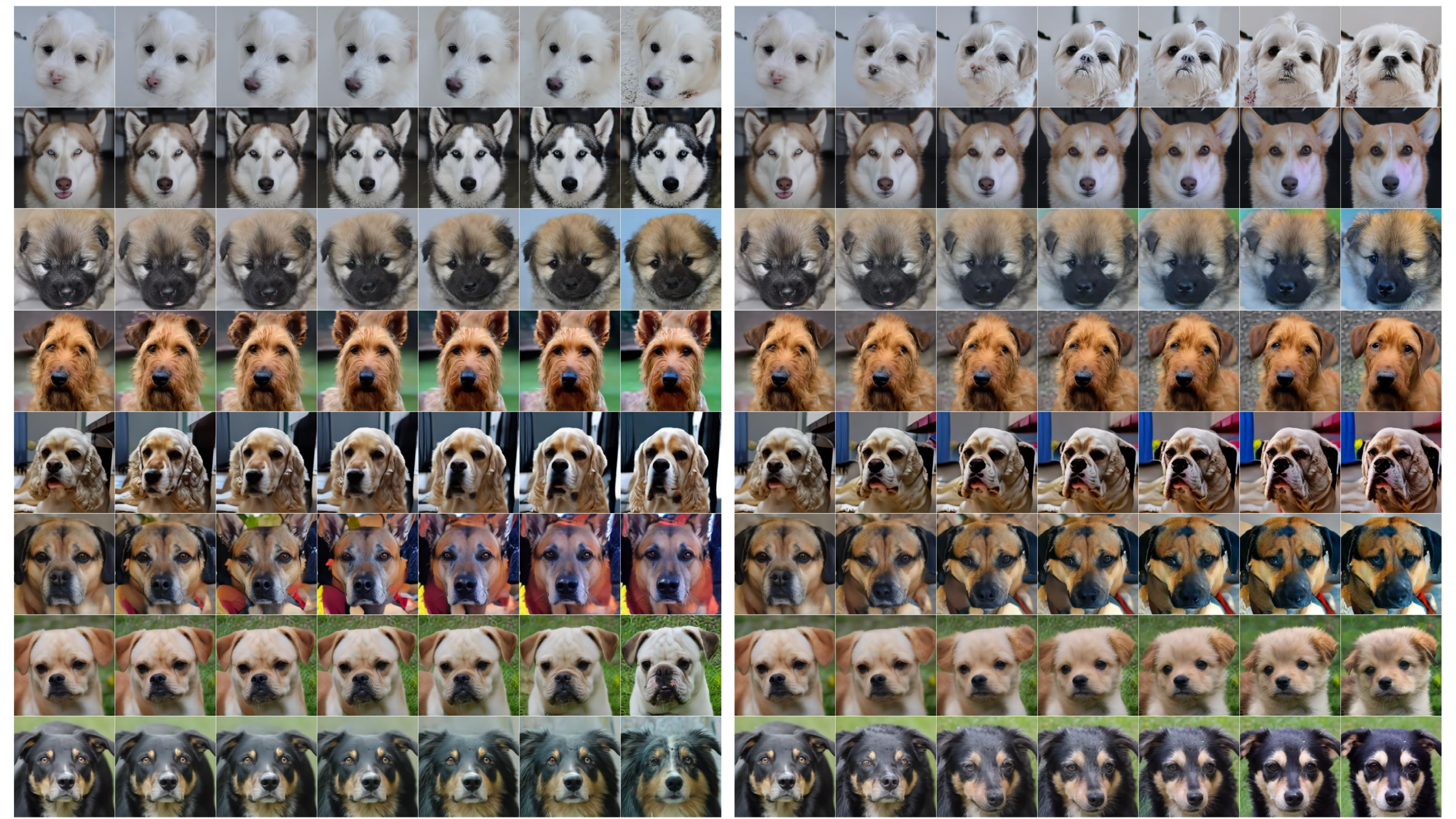}
    \caption{
    \textbf{$\boldsymbol{t=0.75T}$.} A selection of interpretable edits discovered by our feature direction in AFHQ. The image on the far left represents the reconstructed original image, while the subsequent images demonstrate the interpretable edits that have been made to it.}
    \label{fig:afhq_local_appendix_75}
\end{figure}

\begin{figure}[!h]
    \centering
    \includegraphics[width=1.0\linewidth]{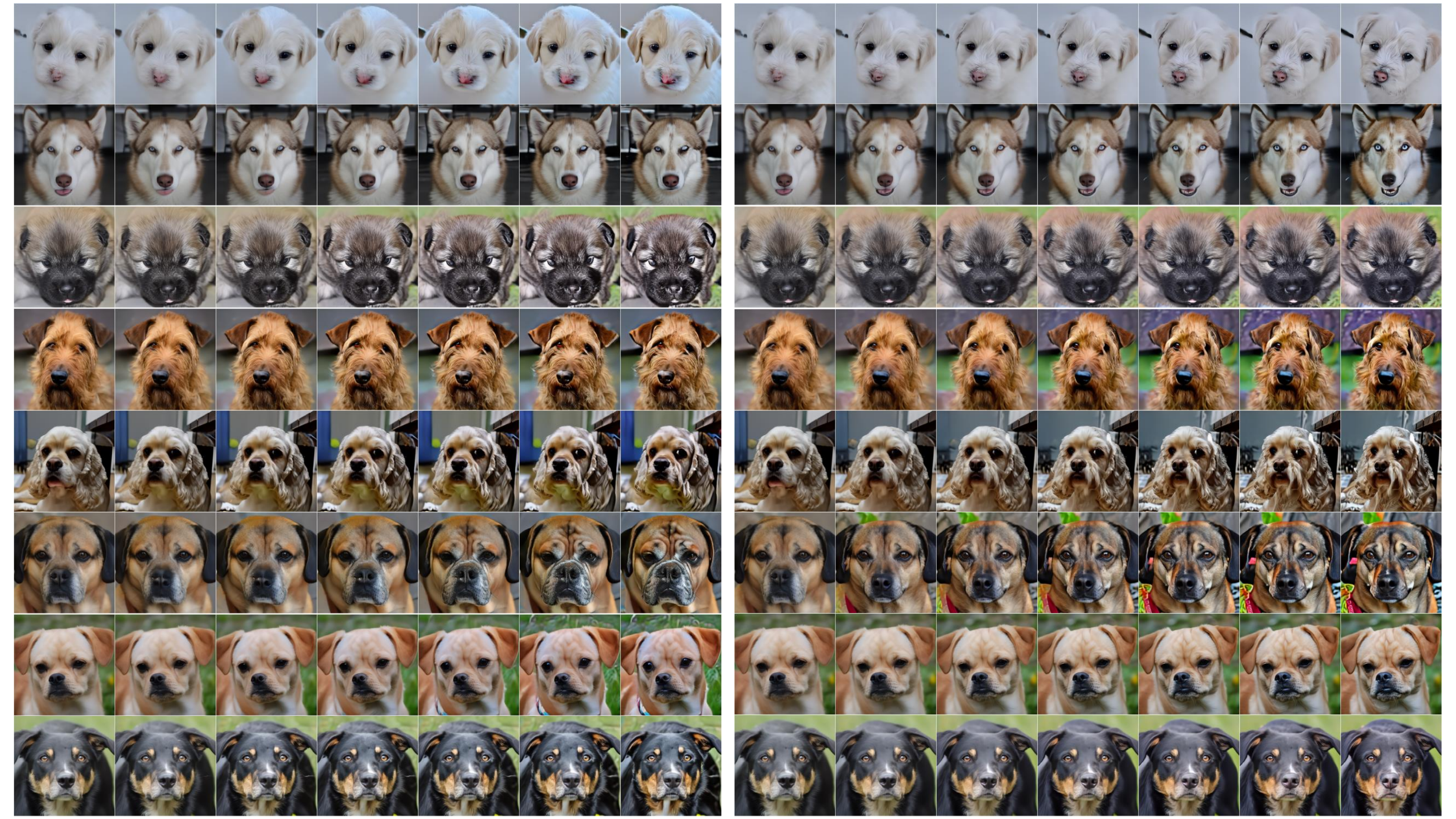}
    \caption{
    \textbf{$\boldsymbol{t=0.5T}$.} A selection of interpretable edits discovered by our feature direction in AFHQ. The image on the far left represents the reconstructed original image, while the subsequent images demonstrate the interpretable edits that have been made to it.}
    \label{fig:afhq_local_appendix_50}
\end{figure}

\begin{figure}[!h]
    \centering
    \includegraphics[width=1.0\linewidth]{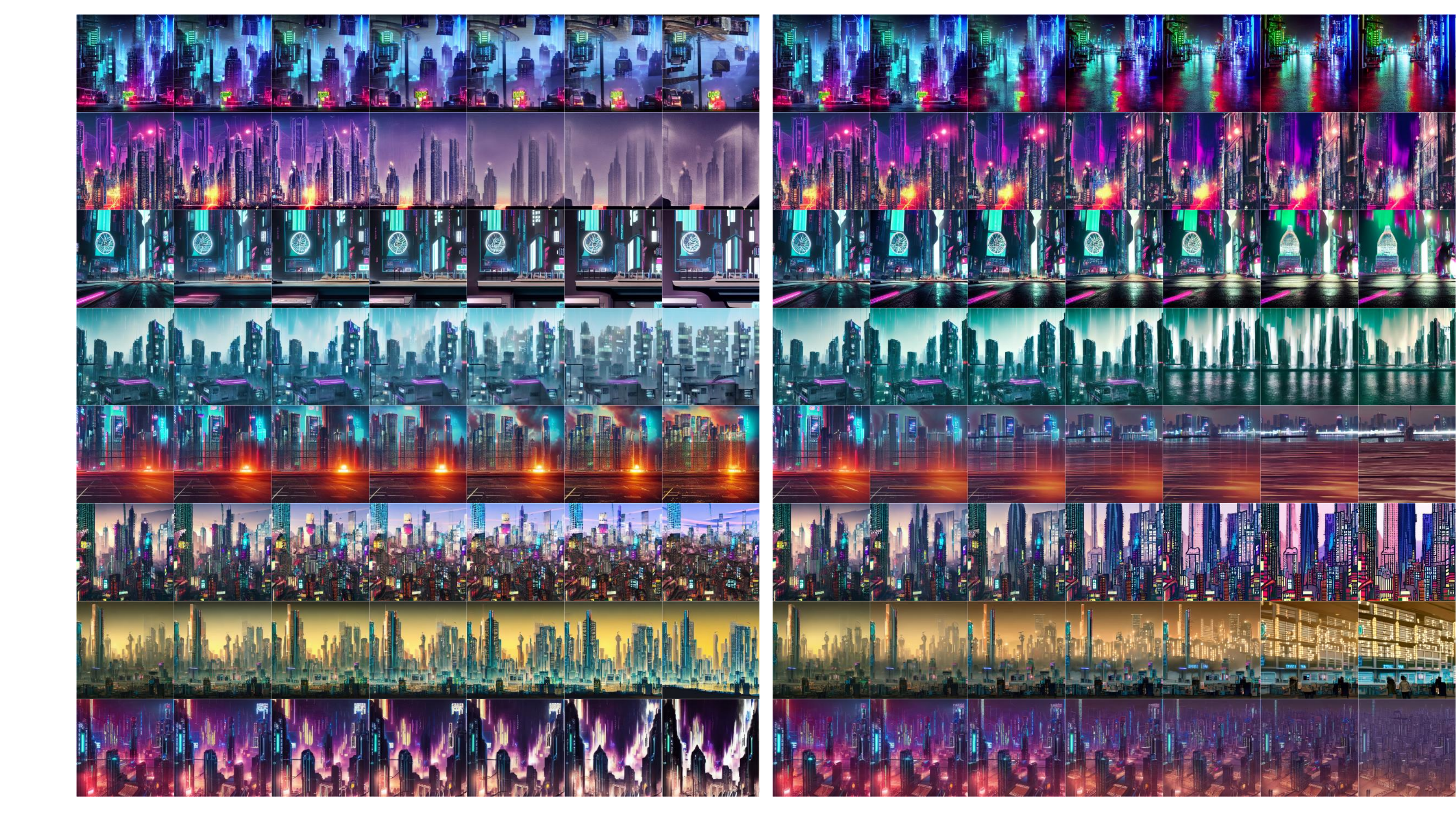}
    \caption{
    \textbf{$\boldsymbol{t=T}$.} A selection of interpretable edits discovered by our feature direction in LDM with "Cyberpunk city". The image on the far left represents the reconstructed original image, while the subsequent images demonstrate the interpretable edits that have been made to it.}
    \label{fig:ldm_cyber_appendix_1T}
\end{figure}

\begin{figure}[!h]
    \centering
    \includegraphics[width=1.0\linewidth]{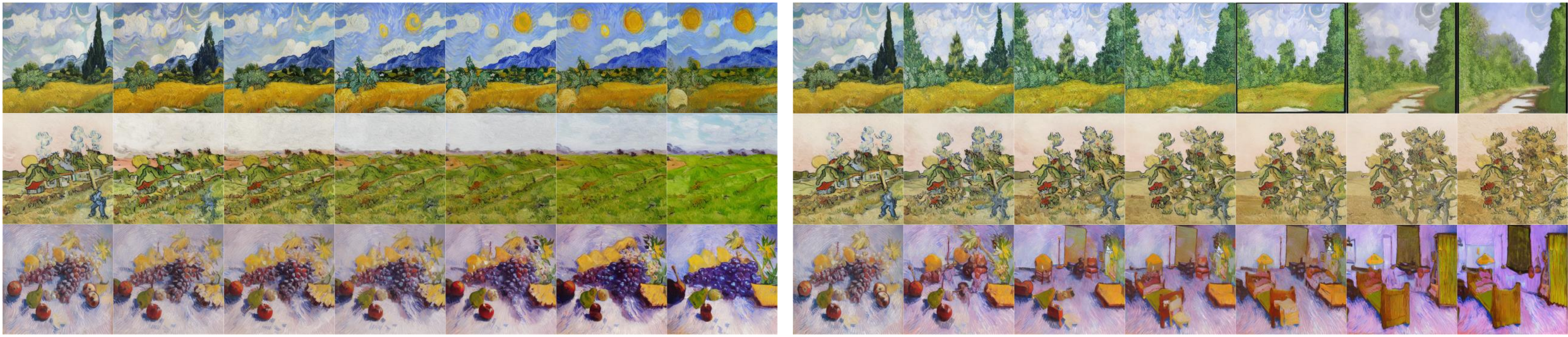}
    \caption{
    \textbf{$\boldsymbol{t=T}$.} A selection of interpretable edits discovered by our feature direction in LDM with "Painting of VanGogh". The image on the far left represents the reconstructed original image, while the subsequent images demonstrate the interpretable edits that have been made to it.}
    \label{fig:ldm_local_gogh_appendix_T1}
\end{figure}

\begin{figure}[!h]
    \centering
    \includegraphics[width=1.0\linewidth]{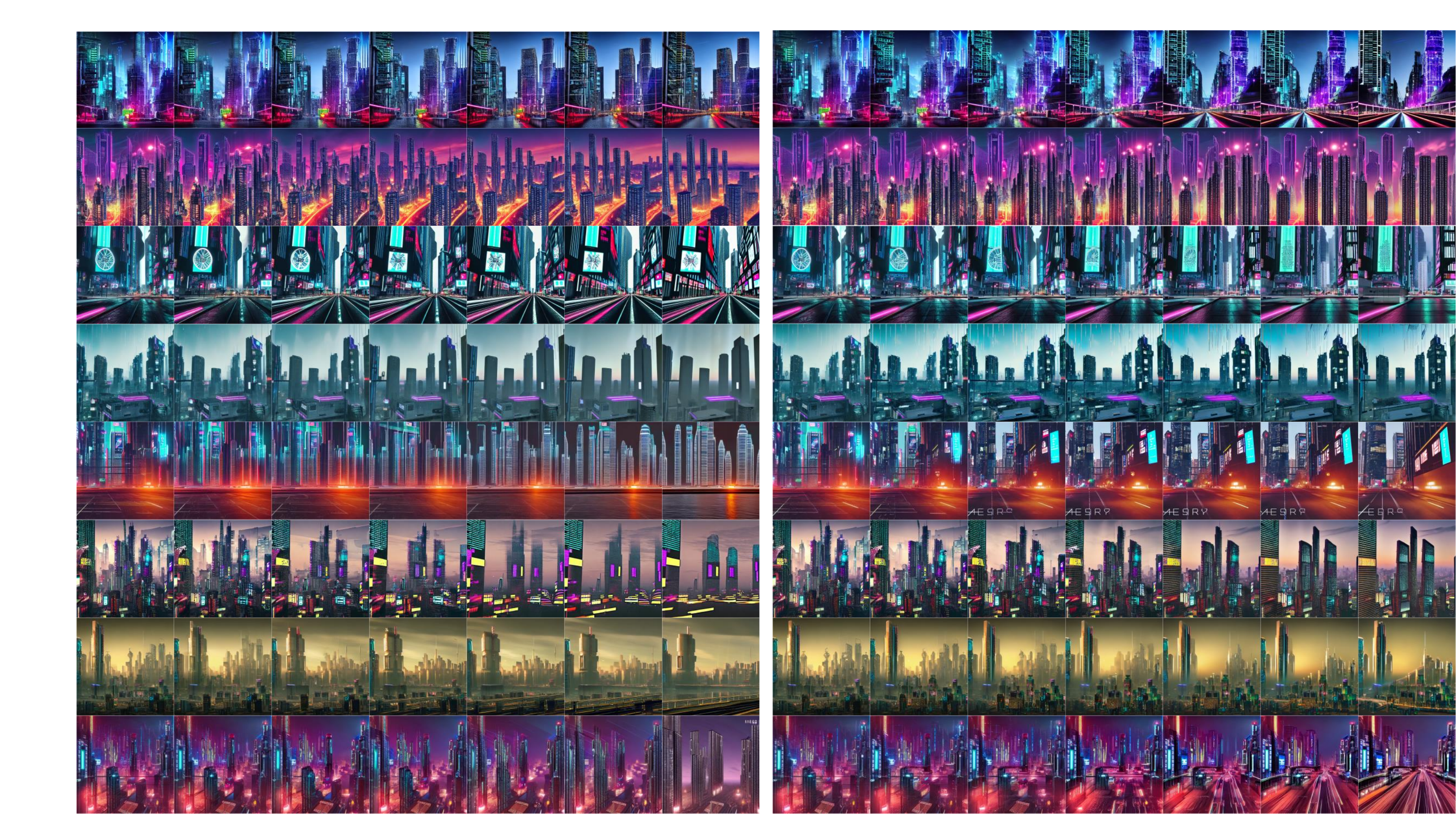}
    \caption{
    \textbf{$\boldsymbol{t=0.75T}$.} A selection of interpretable edits discovered by our feature direction in LDM with "Cyberpunk city". The image on the far left represents the reconstructed original image, while the subsequent images demonstrate the interpretable edits that have been made to it.}
    \label{fig:ldm_cyber_appendix_75T}
\end{figure}

\begin{figure}[!h]
    \centering
    \includegraphics[width=1.0\linewidth]{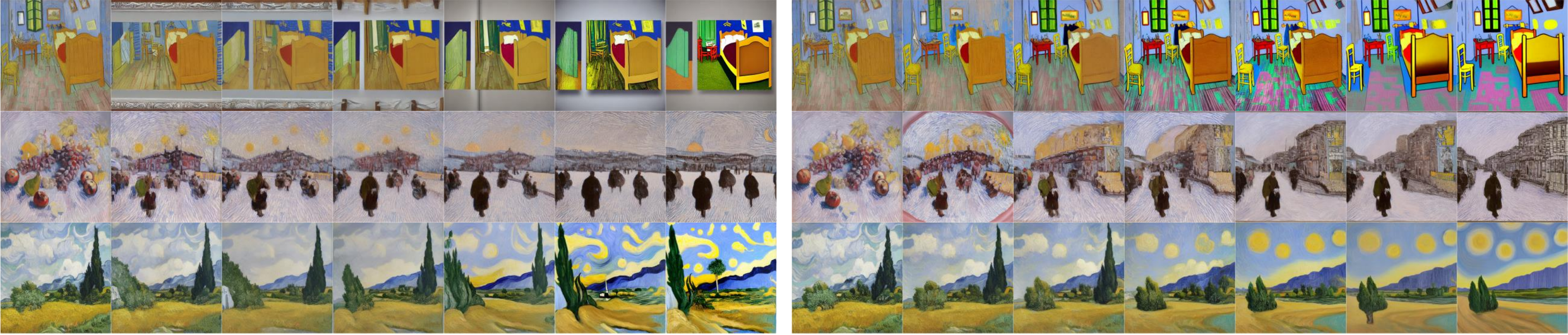}
    \caption{
    \textbf{$\boldsymbol{t=0.75T}$.} A selection of interpretable edits discovered by our feature direction in LDM with "Painting of VanGogh". The image on the far left represents the reconstructed original image, while the subsequent images demonstrate the interpretable edits that have been made to it.}
    \label{fig:ldm_local_gogh_appendix_T75}
\end{figure}

\begin{figure}[!h]
    \centering
    \includegraphics[width=1.0\linewidth]{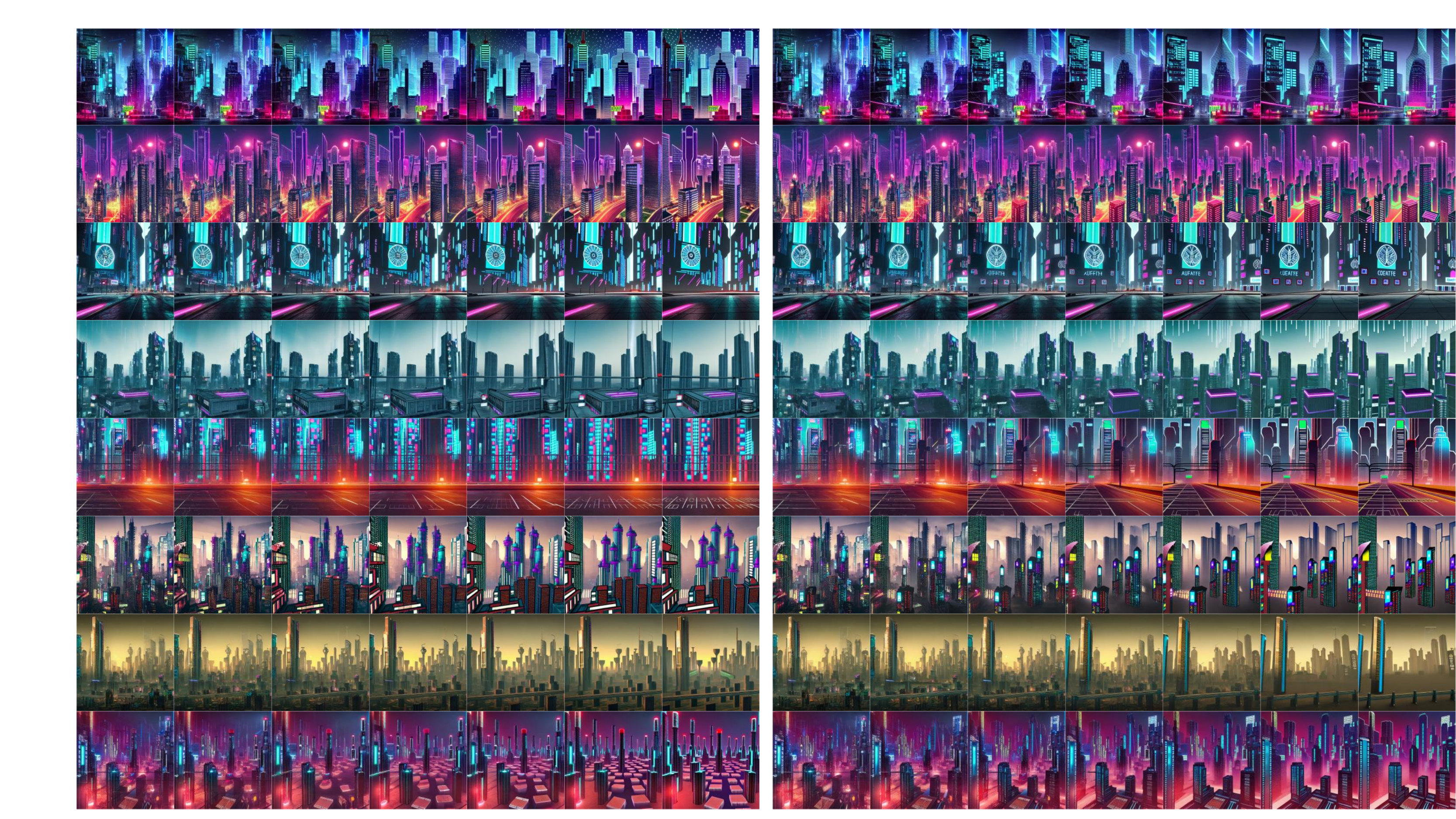}
    \caption{
    \textbf{$\boldsymbol{t=0.5T}$.} A selection of interpretable edits discovered by our feature direction in LDM with "Cyberpunk city". The image on the far left represents the reconstructed original image, while the subsequent images demonstrate the interpretable edits that have been made to it.}
    \label{fig:ldm_cyber_appendix_5T}
\end{figure}

\begin{figure}[!h]
    \centering
    \includegraphics[width=1.0\linewidth]{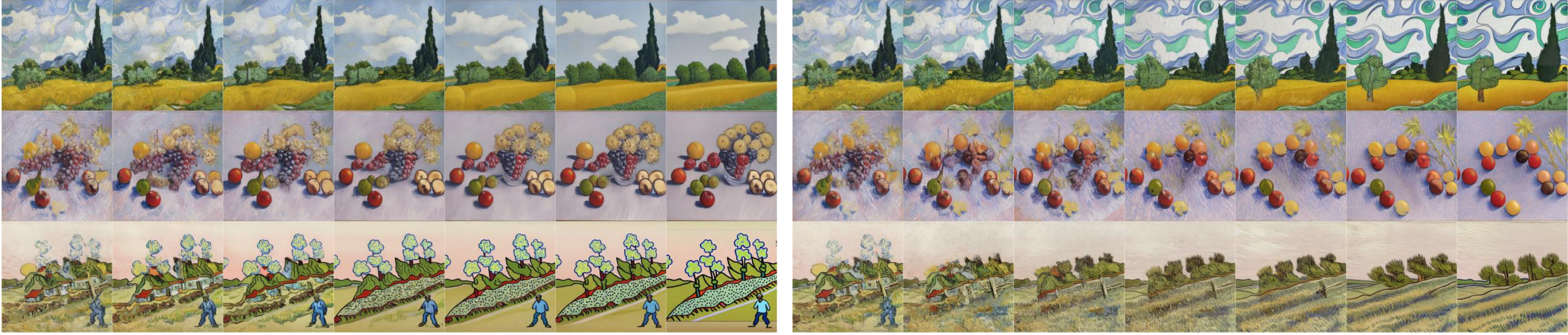}
    \caption{
    \textbf{$\boldsymbol{t=0.5T}$.} A selection of interpretable edits discovered by our feature direction in LDM with "Painting of VanGogh". The image on the far left represents the reconstructed original image, while the subsequent images demonstrate the interpretable edits that have been made to it.}
    \label{fig:ldm_local_gogh_appendix_T5}
\end{figure}

\clearpage
\subsection{Global feature direction}
\label{appendix:global}

\begin{figure}[!h]
    \centering
    \includegraphics[width=1.0\linewidth]{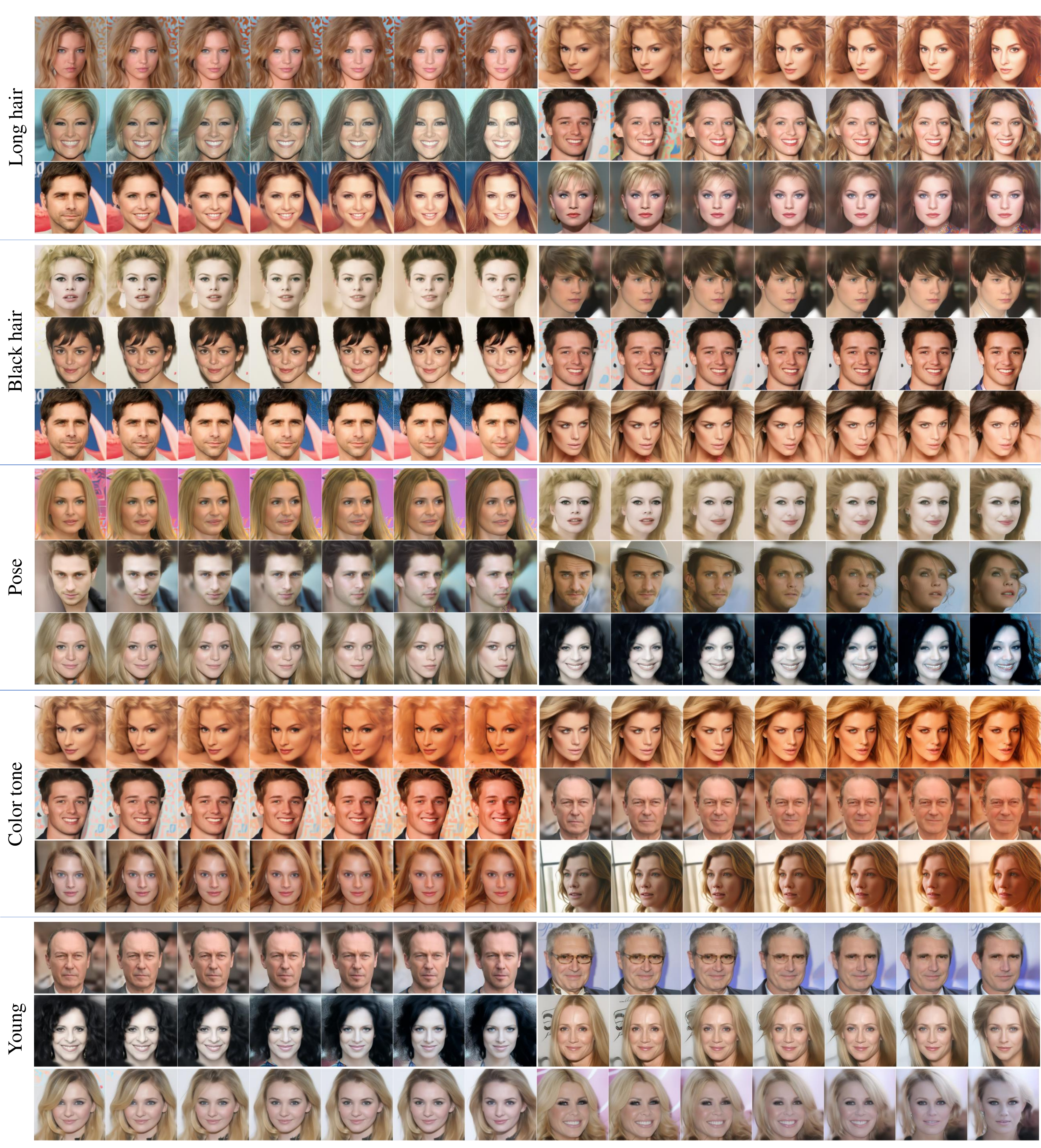}
    \caption{
    A selection of interpretable edits discovered by our global feature direction in CelebA-HQ. The image on the far left represents the reconstructed original image, while the subsequent images demonstrate the interpretable edits that have been made to it.}
    \label{fig:celeba_global_appendix}
\end{figure}

\begin{figure}[!t]
    \centering
    \includegraphics[width=1.0\linewidth]{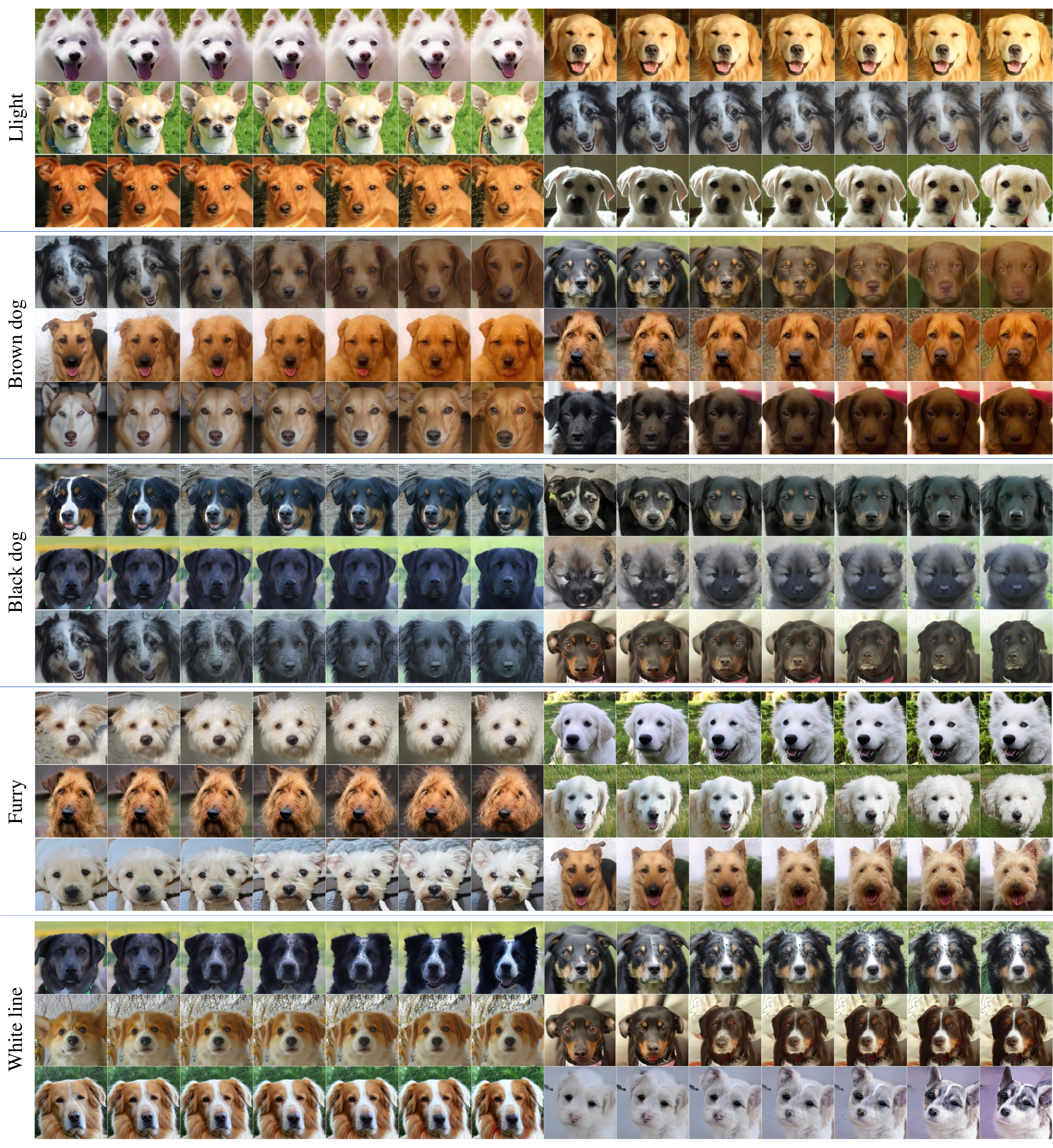}
    \caption{
    A selection of interpretable edits discovered by our global feature direction in AFHQ. The image on the far left represents the reconstructed original image, while the subsequent images demonstrate the interpretable edits that have been made to it.}
    \label{fig:afhq_global_appendix2}
\end{figure}

\begin{figure}[!t]
    \centering
    \includegraphics[width=1.0\linewidth]{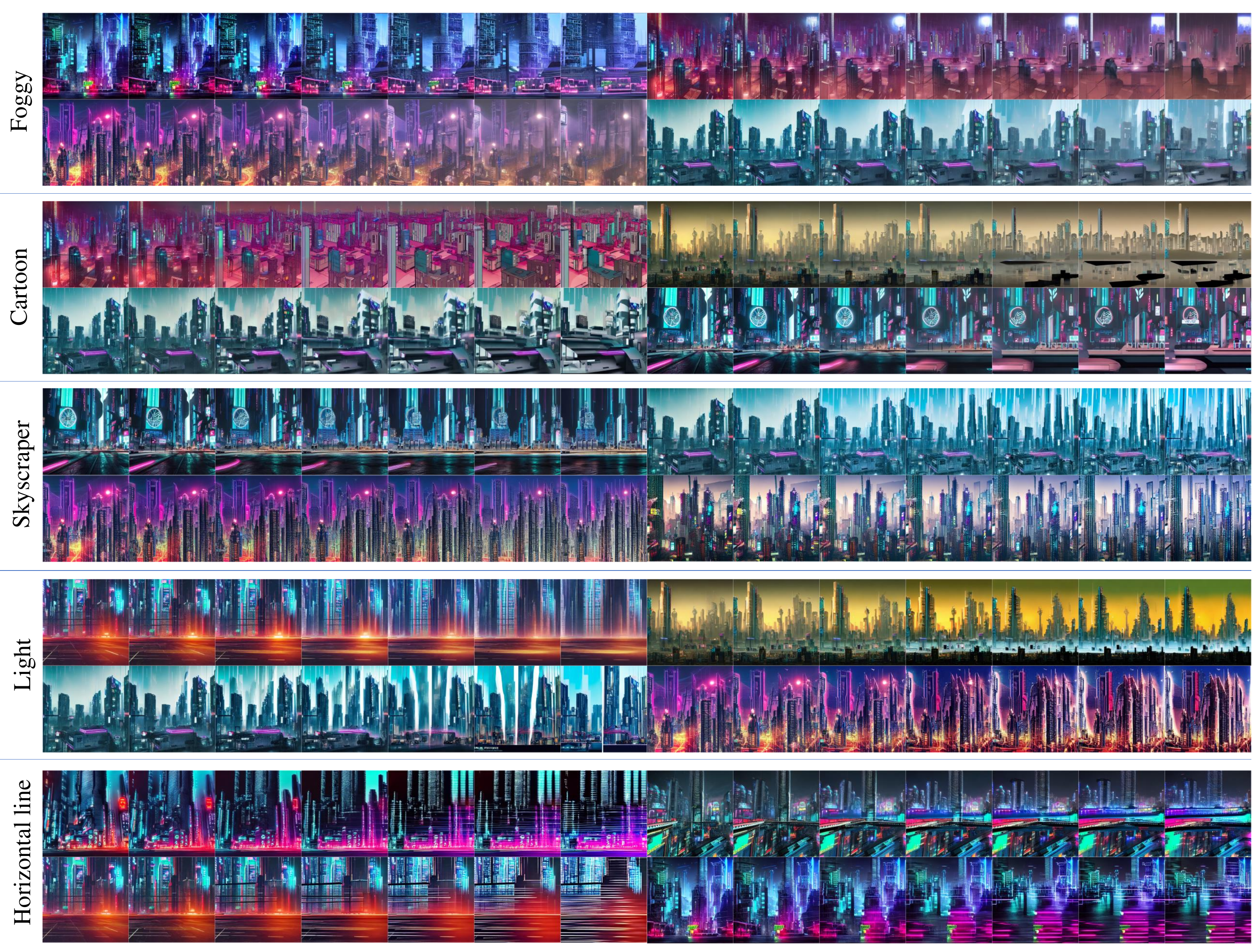}
    \caption{
    A selection of interpretable edits discovered by our global feature direction in LDM with "Cyberpunk city". The image on the far left represents the reconstructed original image, while the subsequent images demonstrate the interpretable edits that have been made to it.}
    \label{fig:LDM_global_appendix}
\end{figure}

%% file: 0main.bbl
\begin{thebibliography}{44}
\providecommand{\natexlab}[1]{#1}
\providecommand{\url}[1]{\texttt{#1}}
\expandafter\ifx\csname urlstyle\endcsname\relax
  \providecommand{\doi}[1]{doi: #1}\else
  \providecommand{\doi}{doi: \begingroup \urlstyle{rm}\Url}\fi

\bibitem[Abdal et~al.(2021)Abdal, Zhu, Mitra, and Wonka]{abdal2021styleflow}
Abdal, R., Zhu, P., Mitra, N.~J., and Wonka, P.
\newblock Styleflow: Attribute-conditioned exploration of stylegan-generated
  images using conditional continuous normalizing flows.
\newblock \emph{ACM Transactions on Graphics (ToG)}, 40\penalty0 (3):\penalty0
  1--21, 2021.

\bibitem[Arvanitidis et~al.(2017)Arvanitidis, Hansen, and
  Hauberg]{arvanitidis2017latent}
Arvanitidis, G., Hansen, L.~K., and Hauberg, S.
\newblock Latent space oddity: on the curvature of deep generative models.
\newblock \emph{arXiv preprint arXiv:1710.11379}, 2017.

\bibitem[Arvanitidis et~al.(2020)Arvanitidis, Hauberg, and
  Sch{\"o}lkopf]{arvanitidis2020geometrically}
Arvanitidis, G., Hauberg, S., and Sch{\"o}lkopf, B.
\newblock Geometrically enriched latent spaces.
\newblock \emph{arXiv preprint arXiv:2008.00565}, 2020.

\bibitem[Avrahami et~al.(2022{\natexlab{a}})Avrahami, Hayes, Gafni, Gupta,
  Taigman, Parikh, Lischinski, Fried, and Yin]{avrahami2022spatext}
Avrahami, O., Hayes, T., Gafni, O., Gupta, S., Taigman, Y., Parikh, D.,
  Lischinski, D., Fried, O., and Yin, X.
\newblock Spatext: Spatio-textual representation for controllable image
  generation.
\newblock \emph{arXiv preprint arXiv:2211.14305}, 2022{\natexlab{a}}.

\bibitem[Avrahami et~al.(2022{\natexlab{b}})Avrahami, Lischinski, and
  Fried]{avrahami2022blended}
Avrahami, O., Lischinski, D., and Fried, O.
\newblock Blended diffusion for text-driven editing of natural images.
\newblock In \emph{Proceedings of the IEEE/CVF Conference on Computer Vision
  and Pattern Recognition}, pp.\  18208--18218, 2022{\natexlab{b}}.

\bibitem[Balaji et~al.(2022)Balaji, Nah, Huang, Vahdat, Song, Kreis, Aittala,
  Aila, Laine, Catanzaro, et~al.]{balaji2022ediffi}
Balaji, Y., Nah, S., Huang, X., Vahdat, A., Song, J., Kreis, K., Aittala, M.,
  Aila, T., Laine, S., Catanzaro, B., et~al.
\newblock ediffi: Text-to-image diffusion models with an ensemble of expert
  denoisers.
\newblock \emph{arXiv preprint arXiv:2211.01324}, 2022.

\bibitem[Baranchuk et~al.(2021)Baranchuk, Rubachev, Voynov, Khrulkov, and
  Babenko]{baranchuk2021label}
Baranchuk, D., Rubachev, I., Voynov, A., Khrulkov, V., and Babenko, A.
\newblock Label-efficient semantic segmentation with diffusion models.
\newblock \emph{arXiv preprint arXiv:2112.03126}, 2021.

\bibitem[Chen et~al.(2018)Chen, Klushyn, Kurle, Jiang, Bayer, and
  Smagt]{chen2018metrics}
Chen, N., Klushyn, A., Kurle, R., Jiang, X., Bayer, J., and Smagt, P.
\newblock Metrics for deep generative models.
\newblock In \emph{International Conference on Artificial Intelligence and
  Statistics}, pp.\  1540--1550. PMLR, 2018.

\bibitem[Choi et~al.(2021{\natexlab{a}})Choi, Kim, Jeong, Gwon, and
  Yoon]{choi2021ilvr}
Choi, J., Kim, S., Jeong, Y., Gwon, Y., and Yoon, S.
\newblock Ilvr: Conditioning method for denoising diffusion probabilistic
  models.
\newblock \emph{arXiv preprint arXiv:2108.02938}, 2021{\natexlab{a}}.

\bibitem[Choi et~al.(2021{\natexlab{b}})Choi, Lee, Yoon, Park, Hwang, and
  Kang]{choi2021not}
Choi, J., Lee, J., Yoon, C., Park, J.~H., Hwang, G., and Kang, M.
\newblock Do not escape from the manifold: Discovering the local coordinates on
  the latent space of gans.
\newblock \emph{arXiv preprint arXiv:2106.06959}, 2021{\natexlab{b}}.

\bibitem[Choi et~al.(2022)Choi, Lee, Shin, Kim, Kim, and
  Yoon]{choi2022perception}
Choi, J., Lee, J., Shin, C., Kim, S., Kim, H., and Yoon, S.
\newblock Perception prioritized training of diffusion models.
\newblock In \emph{Proceedings of the IEEE/CVF Conference on Computer Vision
  and Pattern Recognition}, pp.\  11472--11481, 2022.

\bibitem[Choi et~al.(2018)Choi, Choi, Kim, Ha, Kim, and Choo]{choi2018stargan}
Choi, Y., Choi, M., Kim, M., Ha, J.-W., Kim, S., and Choo, J.
\newblock Stargan: Unified generative adversarial networks for multi-domain
  image-to-image translation.
\newblock In \emph{Proceedings of the IEEE conference on computer vision and
  pattern recognition}, pp.\  8789--8797, 2018.

\bibitem[Daras \& Dimakis(2022)Daras and Dimakis]{daras2022multiresolution}
Daras, G. and Dimakis, A.~G.
\newblock Multiresolution textual inversion.
\newblock \emph{arXiv preprint arXiv:2211.17115}, 2022.

\bibitem[Dhariwal \& Nichol(2021)Dhariwal and Nichol]{dhariwal2021diffusion}
Dhariwal, P. and Nichol, A.
\newblock Diffusion models beat gans on image synthesis.
\newblock \emph{Advances in Neural Information Processing Systems},
  34:\penalty0 8780--8794, 2021.

\bibitem[Goodfellow et~al.(2020)Goodfellow, Pouget-Abadie, Mirza, Xu,
  Warde-Farley, Ozair, Courville, and Bengio]{goodfellow2020generative}
Goodfellow, I., Pouget-Abadie, J., Mirza, M., Xu, B., Warde-Farley, D., Ozair,
  S., Courville, A., and Bengio, Y.
\newblock Generative adversarial networks.
\newblock \emph{Communications of the ACM}, 63\penalty0 (11):\penalty0
  139--144, 2020.

\bibitem[H{\"a}rk{\"o}nen et~al.(2020)H{\"a}rk{\"o}nen, Hertzmann, Lehtinen,
  and Paris]{harkonen2020ganspace}
H{\"a}rk{\"o}nen, E., Hertzmann, A., Lehtinen, J., and Paris, S.
\newblock Ganspace: Discovering interpretable gan controls.
\newblock \emph{Advances in Neural Information Processing Systems},
  33:\penalty0 9841--9850, 2020.

\bibitem[Ho \& Salimans(2022)Ho and Salimans]{ho2022classifier}
Ho, J. and Salimans, T.
\newblock Classifier-free diffusion guidance.
\newblock \emph{arXiv preprint arXiv:2207.12598}, 2022.

\bibitem[Ho et~al.(2020)Ho, Jain, and Abbeel]{ho2020denoising}
Ho, J., Jain, A., and Abbeel, P.
\newblock Denoising diffusion probabilistic models.
\newblock \emph{Advances in Neural Information Processing Systems},
  33:\penalty0 6840--6851, 2020.

\bibitem[Karras et~al.(2018)Karras, Aila, Laine, and
  Lehtinen]{karras2018progressive}
Karras, T., Aila, T., Laine, S., and Lehtinen, J.
\newblock Progressive growing of gans for improved quality, stability, and
  variation.
\newblock In \emph{International Conference on Learning Representations}, 2018.

\bibitem[Karras et~al.(2019)Karras, Laine, and Aila]{karras2019style}
Karras, T., Laine, S., and Aila, T.
\newblock A style-based generator architecture for generative adversarial
  networks.
\newblock In \emph{Proceedings of the IEEE/CVF conference on computer vision
  and pattern recognition}, pp.\  4401--4410, 2019.

\bibitem[Karras et~al.(2022)Karras, Aittala, Aila, and
  Laine]{karras2022elucidating}
Karras, T., Aittala, M., Aila, T., and Laine, S.
\newblock Elucidating the design space of diffusion-based generative models.
\newblock \emph{arXiv preprint arXiv:2206.00364}, 2022.

\bibitem[Kawar et~al.(2022)Kawar, Zada, Lang, Tov, Chang, Dekel, Mosseri, and
  Irani]{kawar2022imagic}
Kawar, B., Zada, S., Lang, O., Tov, O., Chang, H., Dekel, T., Mosseri, I., and
  Irani, M.
\newblock Imagic: Text-based real image editing with diffusion models.
\newblock \emph{arXiv preprint arXiv:2210.09276}, 2022.

\bibitem[Kwon et~al.(2022)Kwon, Jeong, and Uh]{kwon2022diffusion}
Kwon, M., Jeong, J., and Uh, Y.
\newblock Diffusion models already have a semantic latent space.
\newblock \emph{arXiv preprint arXiv:2210.10960}, 2022.

\bibitem[Liew et~al.(2022)Liew, Yan, Zhou, and Feng]{liew2022magicmix}
Liew, J.~H., Yan, H., Zhou, D., and Feng, J.
\newblock Magicmix: Semantic mixing with diffusion models.
\newblock \emph{arXiv preprint arXiv:2210.16056}, 2022.

\bibitem[Liu et~al.(2021)Liu, Park, Azadi, Zhang, Chopikyan, Hu, Shi, Rohrbach,
  and Darrell]{liu2021more}
Liu, X., Park, D.~H., Azadi, S., Zhang, G., Chopikyan, A., Hu, Y., Shi, H.,
  Rohrbach, A., and Darrell, T.
\newblock More control for free! image synthesis with semantic diffusion
  guidance.
\newblock \emph{arXiv preprint arXiv:2112.05744}, 2021.

\bibitem[Meng et~al.(2021)Meng, Song, Song, Wu, Zhu, and Ermon]{meng2021sdedit}
Meng, C., Song, Y., Song, J., Wu, J., Zhu, J.-Y., and Ermon, S.
\newblock Sdedit: Image synthesis and editing with stochastic differential
  equations.
\newblock \emph{arXiv preprint arXiv:2108.01073}, 2021.

\bibitem[Nichol et~al.(2021)Nichol, Dhariwal, Ramesh, Shyam, Mishkin, McGrew,
  Sutskever, and Chen]{nichol2021glide}
Nichol, A., Dhariwal, P., Ramesh, A., Shyam, P., Mishkin, P., McGrew, B.,
  Sutskever, I., and Chen, M.
\newblock Glide: Towards photorealistic image generation and editing with
  text-guided diffusion models.
\newblock \emph{arXiv preprint arXiv:2112.10741}, 2021.

\bibitem[Nichol \& Dhariwal(2021)Nichol and Dhariwal]{nichol2021improved}
Nichol, A.~Q. and Dhariwal, P.
\newblock Improved denoising diffusion probabilistic models.
\newblock In \emph{International Conference on Machine Learning}, pp.\
  8162--8171. PMLR, 2021.

\bibitem[Patashnik et~al.(2021)Patashnik, Wu, Shechtman, Cohen-Or, and
  Lischinski]{patashnik2021styleclip}
Patashnik, O., Wu, Z., Shechtman, E., Cohen-Or, D., and Lischinski, D.
\newblock Styleclip: Text-driven manipulation of stylegan imagery.
\newblock In \emph{Proceedings of the IEEE/CVF International Conference on
  Computer Vision}, pp.\  2085--2094, 2021.

\bibitem[Radford et~al.(2021)Radford, Kim, Hallacy, Ramesh, Goh, Agarwal,
  Sastry, Askell, Mishkin, Clark, et~al.]{radford2021learning}
Radford, A., Kim, J.~W., Hallacy, C., Ramesh, A., Goh, G., Agarwal, S., Sastry,
  G., Askell, A., Mishkin, P., Clark, J., et~al.
\newblock Learning transferable visual models from natural language
  supervision.
\newblock In \emph{International Conference on Machine Learning}, pp.\
  8748--8763. PMLR, 2021.

\bibitem[Ramesh et~al.(2018)Ramesh, Choi, and LeCun]{ramesh2018spectral}
Ramesh, A., Choi, Y., and LeCun, Y.
\newblock A spectral regularizer for unsupervised disentanglement.
\newblock \emph{arXiv preprint arXiv:1812.01161}, 2018.

\bibitem[Ramesh et~al.(2022)Ramesh, Dhariwal, Nichol, Chu, and
  Chen]{ramesh2022hierarchical}
Ramesh, A., Dhariwal, P., Nichol, A., Chu, C., and Chen, M.
\newblock Hierarchical text-conditional image generation with clip latents.
\newblock \emph{arXiv preprint arXiv:2204.06125}, 2022.

\bibitem[Rombach et~al.(2022)Rombach, Blattmann, Lorenz, Esser, and
  Ommer]{rombach2022high}
Rombach, R., Blattmann, A., Lorenz, D., Esser, P., and Ommer, B.
\newblock High-resolution image synthesis with latent diffusion models.
\newblock In \emph{Proceedings of the IEEE/CVF Conference on Computer Vision
  and Pattern Recognition}, pp.\  10684--10695, 2022.

\bibitem[Sehwag et~al.(2022)Sehwag, Hazirbas, Gordo, Ozgenel, and
  Canton]{sehwag2022generating}
Sehwag, V., Hazirbas, C., Gordo, A., Ozgenel, F., and Canton, C.
\newblock Generating high fidelity data from low-density regions using
  diffusion models.
\newblock In \emph{Proceedings of the IEEE/CVF Conference on Computer Vision
  and Pattern Recognition}, pp.\  11492--11501, 2022.

\bibitem[Shao et~al.(2018)Shao, Kumar, and Thomas~Fletcher]{shao2018riemannian}
Shao, H., Kumar, A., and Thomas~Fletcher, P.
\newblock The riemannian geometry of deep generative models.
\newblock In \emph{Proceedings of the IEEE Conference on Computer Vision and
  Pattern Recognition Workshops}, pp.\  315--323, 2018.

\bibitem[Shen \& Zhou(2021)Shen and Zhou]{shen2021closed}
Shen, Y. and Zhou, B.
\newblock Closed-form factorization of latent semantics in gans.
\newblock In \emph{Proceedings of the IEEE/CVF Conference on Computer Vision
  and Pattern Recognition}, pp.\  1532--1540, 2021.

\bibitem[Song et~al.(2020{\natexlab{a}})Song, Meng, and
  Ermon]{song2020denoising}
Song, J., Meng, C., and Ermon, S.
\newblock Denoising diffusion implicit models.
\newblock \emph{arXiv preprint arXiv:2010.02502}, 2020{\natexlab{a}}.

\bibitem[Song et~al.(2020{\natexlab{b}})Song, Sohl-Dickstein, Kingma, Kumar,
  Ermon, and Poole]{song2020score}
Song, Y., Sohl-Dickstein, J., Kingma, D.~P., Kumar, A., Ermon, S., and Poole,
  B.
\newblock Score-based generative modeling through stochastic differential
  equations.
\newblock \emph{arXiv preprint arXiv:2011.13456}, 2020{\natexlab{b}}.

\bibitem[Tumanyan et~al.(2022)Tumanyan, Geyer, Bagon, and
  Dekel]{tumanyan2022plug}
Tumanyan, N., Geyer, M., Bagon, S., and Dekel, T.
\newblock Plug-and-play diffusion features for text-driven image-to-image
  translation.
\newblock \emph{arXiv preprint arXiv:2211.12572}, 2022.

\bibitem[Ye \& Lim(2016)Ye and Lim]{ye2016schubert}
Ye, K. and Lim, L.-H.
\newblock Schubert varieties and distances between subspaces of different
  dimensions.
\newblock \emph{SIAM Journal on Matrix Analysis and Applications}, 37\penalty0
  (3):\penalty0 1176--1197, 2016.

\bibitem[Y{\"u}ksel et~al.(2021)Y{\"u}ksel, Simsar, Er, and
  Yanardag]{yuksel2021latentclr}
Y{\"u}ksel, O.~K., Simsar, E., Er, E.~G., and Yanardag, P.
\newblock Latentclr: A contrastive learning approach for unsupervised discovery
  of interpretable directions.
\newblock In \emph{Proceedings of the IEEE/CVF International Conference on
  Computer Vision}, pp.\  14263--14272, 2021.

\bibitem[Zhang et~al.(2022)Zhang, Tao, and Chen]{zhang2022gddim}
Zhang, Q., Tao, M., and Chen, Y.
\newblock gddim: Generalized denoising diffusion implicit models.
\newblock \emph{arXiv preprint arXiv:2206.05564}, 2022.

\bibitem[Zhang et~al.(2018)Zhang, Isola, Efros, Shechtman, and
  Wang]{zhang2018unreasonable}
Zhang, R., Isola, P., Efros, A.~A., Shechtman, E., and Wang, O.
\newblock The unreasonable effectiveness of deep features as a perceptual
  metric.
\newblock In \emph{Proceedings of the IEEE conference on computer vision and
  pattern recognition}, pp.\  586--595, 2018.

\bibitem[Zhu et~al.(2021)Zhu, Feng, Shen, Zhao, Zha, Zhou, and
  Chen]{zhu2021low}
Zhu, J., Feng, R., Shen, Y., Zhao, D., Zha, Z.-J., Zhou, J., and Chen, Q.
\newblock Low-rank subspaces in gans.
\newblock \emph{Advances in Neural Information Processing Systems},
  34:\penalty0 16648--16658, 2021.

\end{thebibliography}
